\documentclass[journal]{IEEEtran}

\IEEEoverridecommandlockouts
\usepackage{breakurl}
\usepackage{bm}
\usepackage{hyperref}
\usepackage{amsmath}
\usepackage{amssymb}  % assumes amsmath package installed
\let\labelindent\relax

\usepackage{enumitem}
\usepackage{booktabs}
\usepackage{tensor}
\usepackage{amsfonts}
\usepackage{amssymb}
\usepackage{graphicx}
\usepackage{breqn}
\usepackage{siunitx}
\usepackage{afterpage}
\usepackage{multirow}
\usepackage{mathtools}
\usepackage{soul}
\usepackage{tabularx}
\usepackage[table]{xcolor}% http://ctan.org/pkg/xcolor
\definecolor{mygreen}{HTML}{A9D18E}
\definecolor{myblue}{HTML}{9DC3E6}
\newcolumntype{b}{X}
\newcolumntype{s}{>{\hsize=.3\hsize}X}
\newcolumntype{x}{>{\hsize=.5\hsize}X}
\usepackage{tikz}
\usetikzlibrary{shapes.geometric, arrows}
\usepackage{caption}
\usepackage{subcaption}
\usepackage{cite}
\newcommand{\norm}[1]{\left\lVert#1\right\rVert}

\newcommand{\twonorm}[1]{\left\lVert#1\right\rVert_2}

\newcommand{\prths}[1]{\left(#1\right)}

\newcommand{\axis}[2]{\mathbf{e}_{#1}^{#2}}

\newcommand{\id}{k}
\newcommand{\loadf}{\mathcal{L}}
\newcommand{\robotf}{\mathcal{B}}
\newcommand{\worldf}{\mathcal{I}}

\newcommand{\veca}{\boldsymbol{a}}

\newcommand{\vecb}{\mathbf{b}}
\newcommand{\vecc}{\mathbf{c}}

\newcommand{\vece}{\mathbf{e}}
\newcommand{\vecF}{\mathbf{F}}
\newcommand{\vecM}{\mathbf{M}}

\newcommand{\veclamb}{\bm{\Lambda}}

\newcommand{\vecg}{\mathbf{g}}
\newcommand{\vecxi}{\bm{\xi}}

\newcommand{\vecp}{\mathbf{p}}

\newcommand{\vecu}{\mathbf{u}}
\newcommand{\vecmu}{\bm{\mu}}
\newcommand{\vecomega}{\bm{\omega}}
\newcommand{\vecrho}[1]{\bm{\rho}_{#1}}
\newcommand{\hatvecrho}[1]{\hat{\bm{\rho}}_{#1}}

\newcommand{\vecx}{\mathbf{x}}

\newcommand{\mat}[2]{\mathbf{#1}_{#2}}

\newcommand{\matI}{\mathbf{I}}
\newcommand{\matK}{\mathbf{K}}
\newcommand{\matM}{\mathbf{M}}
\newcommand{\matN}{\mathbf{N}}
\newcommand{\matP}{\mathbf{P}}
\newcommand{\matR}{\bm{\mathit{R}}}

\newcommand{\inertia}{\mathbf{J}}
\newcommand{\inertiaload}{\mathbf{J}_{L}}
\newcommand{\loadmass}{m_{L}}
\newcommand{\massi}{m_{\id}}

\newcommand{\half}{\frac{1}{2}}

\newcommand{\angvel}{\mathbf{\Omega}}
\newcommand{\angacc}{\dot{\angvel}}

\newcommand{\robotpos}[1]{\vecx_{#1}}
\newcommand{\robotrot}[1]{\matR_{#1}}

\newcommand{\robotvel}[1]{\dot{\vecx}_{#1}}
\newcommand{\robotacc}[1]{\ddot{\vecx}_{#1}}
\newcommand{\robotangvel}[1]{\angvel_{#1}}
\newcommand{\robotangacc}[1]{\angacc_{#1}}

\newcommand{\tension}[1]{\vecmu_{#1}}
\newcommand{\tensiondes}[1]{\vecmu_{#1,des}}
\newcommand{\tensiondesall}{\vecmu_{des}}
\newcommand{\pseudotensionm}[1]{\vecmu^{0}_{#1}}
\newcommand{\loadpos}{\vecx_{L}}
\newcommand{\loadposdes}{\vecx_{L,des}}
\newcommand{\loadrot}{\matR_{L}}
\newcommand{\loadrotdes}{\matR_{L,des}}

\newcommand{\loadvel}{\dot{\vecx}_{L}}
\newcommand{\loadveldes}{\dot{\vecx}_{L,des}}
\newcommand{\loadacc}{\ddot{\vecx}_{L}}
\newcommand{\loadaccdes}{\ddot{\vecx}_{L,des}}
\newcommand{\loadangvel}{\angvel_{L}}
\newcommand{\loadangveldes}{\angvel_{L,des}}

\newcommand{\loadangacc}{\angacc_{L}}
\newcommand{\loadangaccdes}{\angacc_{L,des}}

\newcommand{\cablevec}[1]{\vecxi_{#1}}

\newcommand{\cableveci}{\cablevec{\id}}
\newcommand{\cablevecides}{\cablevec{\id,des}}
\newcommand{\cablevecnull}[1]{\tilde{\vecxi}_{#1}}
\newcommand{\cablevel}[1]{\vecomega_{#1}}
\newcommand{\cableveldes}[1]{\vecomega_{#1,des}}
\newcommand{\cabledotvec}[1]{\dot{\vecxi}_{#1}}
\newcommand{\cabledotveci}{\cabledotvec{\id}}
\newcommand{\cabledotvecides}{\cabledotvec{\id,des}}
\newcommand{\cableddotvec}[1]{\ddot{\vecxi}_{#1}}

\newcommand{\omegai}{\bm{\omega}_{\id}}

\newcommand{\inputforce}[1]{\vecu_{\mathit{#1}}}
\newcommand{\inputperp}[1]{\vecu_{\mathit{#1}}^{\perp}}
\newcommand{\inputpara}[1]{\vecu_{\mathit{#1}}^{\lVert}}

\newcommand{\ekfinputi}{\mat{U}{k}}
\newcommand{\acci}{\veca_{\mathit{\id}}}
\newcommand{\aptpos}[1]{\vecp_{#1}}
\newcommand{\aptvel}[1]{\dot{\vecp}_{#1}}

\newcommand{\realnum}[1]{\mathbb{R}^{#1}}
\newcommand{\SOt}{SO(3)}
\newcommand{\sot}{\mathfrak{so}(3)}

\newcommand{\extplF}{\mathbf{F}_H}
\newcommand{\extplM}{\mathbf{M}_H}

\newcommand{\scalartension}[1]{\mu_{#1}}

\newcommand{\qdF}[1]{\mathbf{F}_{#1}}
\newcommand{\qdM}[1]{\mathbf{M}_{#1}}

\newcommand{\plFd}{\vecF_{L,des}}
\newcommand{\plMd}{\vecM_{L,des}}

\newcommand{\Phuman}{{\mathbf{p}_{H}}}
\newcommand{\Pobject}{{\mathbf{p}_{O}}}

\newcommand{\netplforce}{\mathbf{F}_L}

\newcommand{\netplmoment}{\mathbf{M}_L}
\newcommand{\patt}[1]{\bold{p}_{att, #1}}

\newcommand{\nullspacev}[1]{\tilde{\bm{\mu}}_{#1}}

\newcommand{\nullproj}{\mathbf{B}}
\newcommand{\softlimit}{\mathbf{Q}}

\newcommand{\robotlimit}{\prescript{\text{r}}{}{r}}
\newcommand{\humanlimit}{\prescript{\text{h}}{}{r}}
\newcommand{\robotypr}[1]{\mathbf{\Theta}_{#1}}
\newcommand{\state}[1]{\mathbf{S}_{#1}}

\newcommand{\controlin}[1]{\mathbf{U}_{#1}}
\newcommand{\measurement}[1]{\mathbf{Z}_{#1}}
\newcommand{\statebar}[1]{\bar{\mathbf{S}}_{#1}}

\newcommand{\kalmangain}[1]{\mathbf{K}_{#1}}
\newcommand{\measurementmodel}{\mathbf{H}}

\usepackage{wasysym}  % symbols
\author{
Guanrui Li$^*$, Xinyang Liu$^*$, and Giuseppe Loianno
\thanks{$^*$These authors contributed equally.}
\thanks{The authors are with the New York University, Tandon School of Engineering, Brooklyn, NY 11201, USA. {\tt\footnotesize email: \{lguanrui, liuxy, loiannog\}@nyu.edu}.}
\thanks{This work was supported by the NSF CPS Grant CNS-2121391, the NSF CAREER Award 2145277, Qualcomm Research, Nokia, and NYU Wireless.}
\thanks{The authors acknowledge Jimmy Lee, Manling Li, and Luca Morando for their help and support during the experiments.}
}

\title{\LARGE \bf Human-Aware Physical Human-Robot Collaborative Transportation and Manipulation with Multiple Aerial Robots}

\begin{document}

\markboth{IEEE Transactions on Robotics, 2024}%
{Shell \MakeLowercase{\textit{et al.}}: Bare Demo of IEEEtran.cls for IEEE Journals}

\makeatletter
\g@addto@macro\@maketitle{
\setcounter{figure}{0}
\centering
    \includegraphics[width=\textwidth]{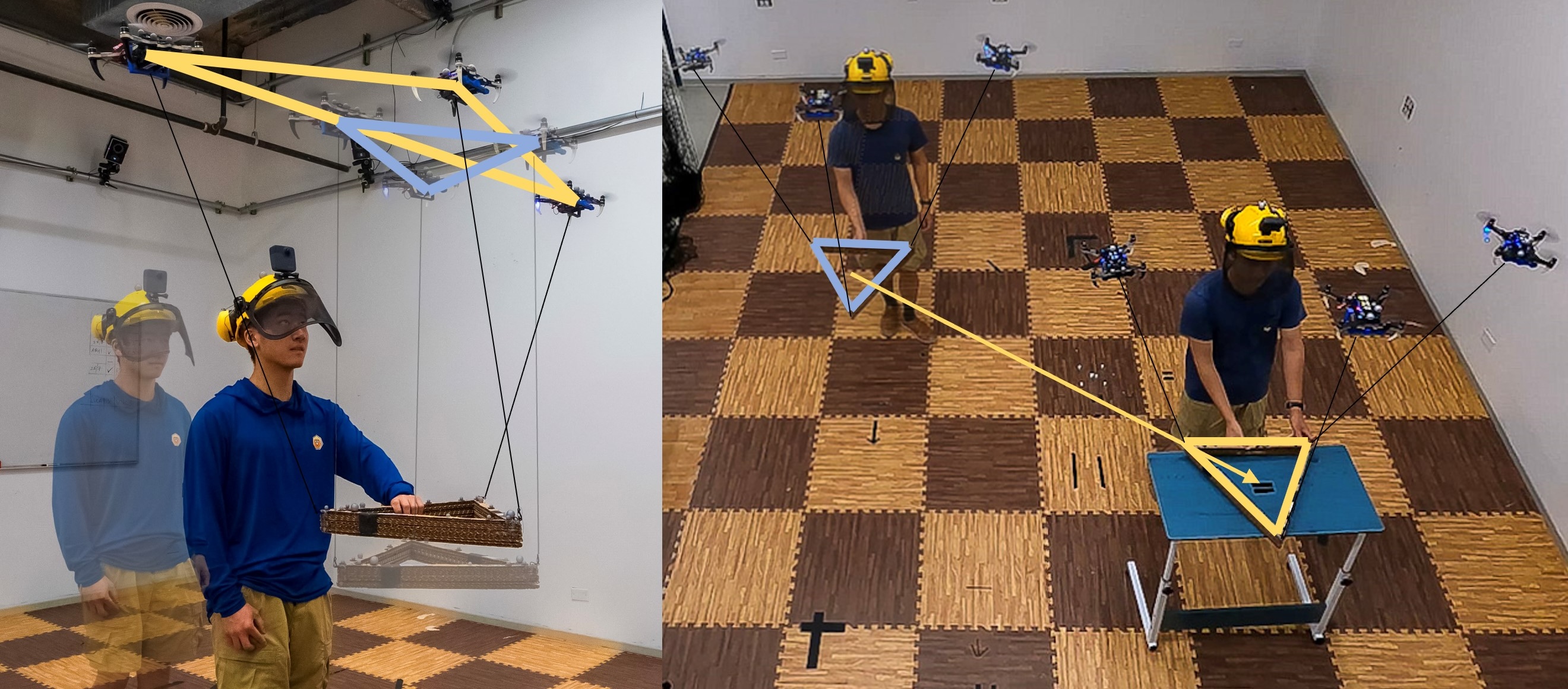}
	\captionof{figure}{A human operator collaborates with three quadrotors, transporting and manipulating a payload. \textit{On the left:} The quadrotors are moving to keep a distance from the human operator without affecting payload tracking. \textit{On the right:} the aerial robot team is transporting the payload using the human operator's interactive force and moment as commands.\label{fig:Intro}}
	\vspace{-3ex}
}
\makeatother

\maketitle

\begin{abstract}
Human-robot interaction will play an essential role in various industries and daily tasks, enabling robots to effectively collaborate with humans and reduce their physical workload. Most of the existing approaches for physical human-robot interaction focus on collaboration between a human and a single ground or aerial robot. In recent years, very little progress has been made in this research area when considering multiple aerial robots, which offer increased versatility and mobility. This paper proposes a novel approach for physical human-robot collaborative transportation and manipulation of a cable-suspended payload with multiple aerial robots. The proposed method enables smooth and intuitive interaction between the transported objects and a human worker. In the same time, we consider distance constraints during the operations by exploiting the internal redundancy of the multi-robot transportation system. The key elements of our approach are (a) a collaborative payload external wrench estimator that does not rely on any force sensor; (b) a 6D admittance controller for human-aerial-robot collaborative transportation and manipulation; (c) a human-aware force distribution that exploits the internal system redundancy to guarantee the execution of additional tasks such inter-human-robot separation without affecting the payload trajectory tracking or quality of interaction. We validate the approach through extensive simulation and real-world experiments. These include scenarios where the robot team assists the human in transporting and manipulating a load, or where the human helps the robot team navigate the environment. We experimentally demonstrate for the first time, to the best of our knowledge, that our approach enables a quadrotor team to physically collaborate with a human in manipulating a payload in all 6 DoF in collaborative human-robot transportation and manipulation tasks.
\end{abstract}

\begin{IEEEkeywords}
Aerial Robotics, Physical Human-Robot Interaction
\end{IEEEkeywords}

\IEEEpeerreviewmaketitle
\section*{Supplementary material}
\textbf{Video}: \url{https://youtu.be/WMev7j97fDg}

\section{Introduction}\label{sec:intro}
As envisioned in the Industry 4.0 revolution, human-robot interaction will play an increasingly significant role in future industries and daily life~\cite{Schwab_2017}. While most research in human-robot interaction has concentrated on collaborations between humans and ground robots, only a limited number of approaches have been developed for aerial robots, with the majority being confined to teleoperation. Unlike ground robots, collaborative Micro Aerial Vehicles (MAVs) show additional flexibility and maneuverability due to their 3D mobility and compact size. Moreover, a team of collaborative MAVs can provide increased adaptivity, resilience, and robustness during a task or multiple simultaneous tasks compared to a single aerial robot. For example, MAV teams can assist humans in executing complex or dangerous tasks, including but not limited to inspection~\cite{mansouriCooperativeCoveragePath2018,shah2020multidronesurveys},
mapping~\cite{giuseppe2018ijrr,michael2012jfrcollaborativemapping}, environment interaction~\cite{cano2022jirs,sanalitro2022ralindirectforce}, surveillance~\cite{saskaSwarmDistributionDeployment2016}, and autonomous transportation and manipulation~\cite{guanrui2021iser,jackson2020raldistributedplanning}. Specifically, in autonomous aerial transportation and manipulation, there are many possible usage scenarios. For instance, in a post-disaster response task, a team of aerial robots can cooperatively deliver emergency supplies to designated rescue locations based on the first respondent's guidance. Alternatively, on construction sites, an aerial robot team can cooperatively manipulate over-sized construction materials with human workers to expedite the installation process and reduce physical workload.

This paper proposes a novel approach that enables a team of aerial robots to transport and manipulate a cable-suspended payload in physical collaboration with a human operator, as depicted in Fig.~\ref{fig:Intro}.
As discussed in~\cite{guanrui2021ral}, cable mechanisms stand out compared to other existing solutions, such as simple spherical joints, or robot arms~\cite{ollero2021TRO,afifi2023icuas}, because of their lighter weight, lower costs, simpler design requirements, and zero-actuation-energy consumption. Therefore, they are particularly suited for Size, Weight, and Power (SWaP) aerial platforms.

Cables also present a good balance among maneuverability, manipulability, and safety for physical human-aerial-robots collaboration compared to other solutions.
For instance, several solutions attach the robots directly to the payloads via passive mechanisms like spherical joints ~\cite{tagliabue2019ijrr}, magnets~\cite{LoiannoRAL2018} or active mechanisms like grippers~\cite{Mellinger2013}. However, these mechanisms offer reduced maneuverability and manipulability during a manipulation or physical interaction task compared to cables. Conversely, other complex actuated solutions based on robot arms~\cite{thomas2014toward,afifi2022icraphysicalinteractionaerialmanipulator,corsini2022irosnmpchandover} can enhance maneuverability and flexibility. However, this generally comes at the price of increased system inertia and power, potentially compromising the operator's safety. Therefore, compared to other existing solutions, we believe that lightweight cable mechanisms can provide a good trade-off in terms of maneuverability, manipulability, and safety while concurrently offering good flexibility to execute multiple tasks.

We present an innovative control, planning, and estimation framework that enables a human operator to physically collaborate with a team of quadrotors for the transportation and manipulation of a rigid-body payload in all 6 Degrees of Freedom (DoF). A key contribution of this work is the exploitation of system redundancy, allowing for secondary tasks, such as human-aware human-robot interaction. Specifically, our approach ensures distancing between agents and the human operator during physical collaboration, enabling effective human-aware interaction, as depicted in Fig.~\ref{fig:Intro} (left).

Existing approaches to human and aerial robot collaboration have largely focused on single aerial robot interactions~\cite{afifi2022icraphysicalinteractionaerialmanipulator,tognon2021trophysicalinteractiontether,kondak2020icrateleaerialmanipulator}. When considering multiple aerial robots, teleoperation becomes a common solution~\cite{cano2022jirs,gengJGCD2020}. However, few solutions exist for human physical interaction and collaboration with several MAVs~\cite{rastgoftar2019tcst,romano2022jgcd}. However, the human operator's physical collaboration is limited to a 2D horizontal plane. Moreover, these approaches overlook the potential of exploiting the additional DoF to enhance both the system's awareness of the human's presence.

In summary, the contributions of this paper are the following
\begin{itemize}
\item We propose a novel control method that enables a team of quadrotors to manipulate a payload while exploiting the system's redundancy to achieve secondary tasks (e.g., maintaining distances from the human operator or ensuring inter-robot separation). This solution facilitates the physical interaction between the quadrotor team and a human operator.
\item We introduce a collaborative external wrench estimator that allows the robot team to collaboratively measure an external human force input  without relying on any external force sensors. Additionally, we demonstrate that this approach outperforms existing state-of-the-art solutions.
\item We complement our control solution with a 6-DoF admittance controller, which utilizes the estimated human wrench. It enables physical interaction between a human operator and a team of aerial robots for collaborative manipulation and transportation tasks.
\item We experimentally demonstrate for the first time, to the best of our knowledge, that our approach enables a quadrotor team to physically collaborate with a human in manipulating a payload in all 6 DoF in collaborative human-robot transportation and manipulation tasks.
\end{itemize}

The remainder of the paper is organized as follows. In Section~\ref{sec:related_works}, we review relevant literature on cooperative aerial manipulation and physical human-robot interaction. In Section~\ref{sec:Dynamics}, we review the nonlinear system dynamics, considering the external wrench from a human operator. In Section~\ref{sec:Control}, we discuss the proposed human-aware control framework that considers the nonlinear system dynamics. Section~\ref{sec:admittance} details the state estimation strategy and admittance control framework for intended human-aerial-robot collaborative manipulation. Section~\ref{sec:experimental_results} presents real-world experiment results validating the proposed framework. Section~\ref{sec:conclusion} concludes the work and proposes multiple future research directions. 

\section{Related Works}\label{sec:related_works}
\subsection{Cooperative Aerial Manipulation}
In the subsequent discussion, we focus on the existing related works on control, planning, and estimation techniques for aerial transportation and manipulation using suspended cables. This focus arises from the distinct advantages that cable mechanisms offer over other methods, as previously mentioned.

Past literature includes several control and estimation methods~\cite{guanrui2021ral,klausen2020passiveoutdoortransportation,gengJGCD2020,Bernard2011jfr} for autonomous aerial transportation and manipulation using multiple aerial robots equipped with cables. For example, several works~\cite{klausen2020passiveoutdoortransportation,Bernard2011jfr,tran2020itiianntransportation,zhang2021ralselftriggedtransportation,LIU2021106673} propose formation controllers for a team of MAVs to fly in a desired formation when carrying the suspended-payload. The carried payload is not modeled as an integrated part of the system but as an external disturbance that each MAV controller tries to compensate for. Therefore, it is expected that these solutions can struggle to transport the payload to a given position. In~\cite{lee2013cdccooperativepointmass}, by assuming the payload is a point-mass, the authors analyze the full nonlinear dynamics of the system. Based on the dynamic model, the authors design a geometric controller to transport the payload to the desired position moving the quadrotor team to accommodate the desired load motions. 
Moreover, the previously mentioned methods~\cite{lee2013cdccooperativepointmass,Bernard2011jfr,tran2020itiianntransportation,zhang2021ralselftriggedtransportation}, treat the payload as a point-mass, hence restricting the manipulation capabilities to payload's positional movements only, with no control over its orientation.

Other approaches for autonomous aerial transportation and manipulation rely on a leader-follower paradigm~\cite{tagliabue2017icracollaborativetransportation,gassner2017icradynamiccollaboration,rastgoftar2018icuas,heng2022access, tognon2018passive}. The leader robot follows the desired trajectory, whereas the followers maintain either a constant distance from the leader~\cite{rastgoftar2018icuas}, or adapt to forces exerted on them when tracking their trajectory~\cite{tagliabue2019ijrr,tagliabue2017icracollaborativetransportation}. However, these methods are subject to error compounding failure since they rely on the leader as a fundamental control unit for navigation. Furthermore, they cannot accurately guarantee the payload's transportation to the desired location or manipulate its orientation.

Several works analyze the system's complex nonlinear dynamics and mechanics and propose corresponding controllers to control the payload's pose in 6 DoF~\cite{lee2018tcst, wu2014geometric,fink2011ijrr,Michael2011}. For example, in~\cite{fink2011ijrr,Michael2011}, the authors assume the system is in a quasi-static state and analyze the corresponding static system mechanics. A payload pose controller assigns the quadrotors' desired position to manipulate the payload to the desired pose. In~\cite{wu2014geometric,lee2018tcst}, the complex nonlinear dynamics in the system are thoroughly analyzed using Lagrangian mechanics. Leveraging this model, nonlinear geometric controllers enable the payload to follow the desired pose trajectory. More recently, several solutions propose optimal control strategies~\cite{sundin2022icradecentralizedmpctransportation,tartaglione2017rasModelPredictiveControl}. Although all these works consider the payload a rigid body, the redundant control DoF available in the system~\cite{Leecdc2014} are not exploited to accommodate additional tasks like obstacle avoidance or to ensure safety distance among agents.

Some recent literature starts to investigate this aspect~\cite{gengJGCD2020,geng2022jais,sanalitro2022ralindirectforce}. However, they are specifically designed for a team of three or four quadrotors. In~\cite{masone2016iros}, the authors attempt to leverage system redundancy for a team with an arbitrary number of aerial robots, implementing an optimization formulation for the parallel robot that exclusively optimizes the tension magnitude in the cables, without considering cable directions. However, the methodology presented in~\cite{masone2016iros} does not provide clear instructions for determining cable direction. Consequently, its applicability for secondary tasks such as obstacle avoidance or spatial separation from the human operator remains ambiguous. In this work, we formulate a human-aware controller for any $n\geq3$ quadrotors that exploit the additional system redundancy at the control level, allowing the system to achieve some secondary tasks, such as avoiding obstacles, inter-robot separation or keeping a safe distance among robots and human operators. 

However, since the above methods are designed to control the payload's pose explicitly, it is essential to have a reasonable estimation of the payload's states (i.e., pose and twists) to be fed back into the controller to have a good tracking performance of the payload's pose. Some estimation approaches can recover the payload pose in~\cite{gengJGCD2020,bulka2022outdoortransportation}, but they rely on GPS and, therefore, cannot be employed in indoor environments or areas where the GPS signal is shadowed. Conversely, in our previous work~\cite{guanrui2021ral}, we tackle the payload pose inference problem using onboard vision sensors and IMU to obtain closed-loop control of the payload pose. 

However, the proposed vision-based estimation method might be subject to onboard visual-inertial odometry drift, leading to some constant offset errors. Although such errors will not affect the stability of the system, they will influence the task performance. For instance, consider a scenario where the system is required to transport a payload to a specific location, but due to state estimation drift, there is a 1-meter offset from the desired destination. In such cases, as shown in this paper, our proposed method enables a local human operator to guide the system and correct the payload to the desired pose, ensuring that the payload is transported to the correct final destination. 

\subsection{Physical Human-Robot Interaction}
Physical Human-Robot Interaction (pHRI) is a rapidly growing field in robotics, facilitating the collaboration between humans and robots in various scenarios such as manufacturing, healthcare, and service industry. 
Most of the research in this field focuses on the collaboration between a single robot and a human.
For instance, this includes cooperative manipulation with a robot arm attached to a ground wheeled base~\cite{spaa2020icra,Stuckler_Behnke_2011}, a fixed-base robot arm~\cite{gienger2018iros}, a humanoid robot~\cite{Sheng_Thobbi_Gu_2015}, or an aerial robot with a manipulator~\cite{afifi2022icraphysicalinteractionaerialmanipulator,tognon2021trophysicalinteractiontether}. 
Past Researchers' works also present approaches like 
admittance control~\cite{Augugliaro_2013_ecc,tognon2021trophysicalinteractiontether}, compliance control, impedance control for single-human-single-robot physical collaboration. 

To increase payload capacity, the common strategies are either deploying a more powerful robot or utilizing a team of robots. The latter approach not only increases redundancy and provides the potential for fault tolerance, but also brings in specialized capabilities that enhance the team's resilience and performance compared to a single robot. However, employing a team of robots demands effective coordination and collaboration across estimation, planning, and control levels.

While multiple robot-human interactions with teleoperation via haptic devices~\cite{SieberIROS2015} or mixed reality glasses~\cite{Yashin_Trinitatova_Agishev_Ibrahimov_Tsetserukou_2019,Sachidanandam_Honarvar_Diaz-Mercado_2022} is a widely explored research topic, physical human-multi-robot cooperative manipulation remains mostly underexplored~\cite{Elwin_Strong_Freeman_Lynch_2023}.
Recently, a handful of works started to research on direct physical interaction between a human and multiple ground robots to cooperatively manipulate or transport an object~\cite{Elwin_Strong_Freeman_Lynch_2023,Sirintuna_Ozdamar_Ajoudani_2023,Carey_Werfel_2022}. In~\cite{Carey_Werfel_2022}, the authors demonstrate that $4$ omnidirectional ground robots with robot arms can physically collaborate with a human using their proposed force-mediated controller but only simulation results were presented.  In~\cite{Elwin_Strong_Freeman_Lynch_2023}, the authors design an omni-robot system that can lift payloads. Then the multiple proposed omni-robots can collaborate with humans to manipulate objects toward desired locations. In~\cite{Sirintuna_Ozdamar_Ajoudani_2023}, the authors use two robot arms with omnidirectional wheeled base to physically collaborate with a human to manipulate over-size objects. They introduce an admittance control module on both ground robots to adapt the human motion as the human leads the manipulation. However, the system dynamics for aerial robots are inherently different from ground robots as aerial robots move in 3D. Hence, the established dynamics models and corresponding control methods for ground robots cannot be directly translated to aerial robots. Moreover, for the SWAP-constrained aerial robots, designing a lightweight, computationally efficient sensing strategy is also essential. 

In \cite{rastgoftar2019tcst}, the authors propose a framework for physical human-robot collaborative transportation of cable-suspended payload with a team of quadrotors. 
The proposed approach models the payload as a point mass and assumes external forces applied on the payload to be constrained in 2D. 
Leveraging their previous work~\cite{rastgoftar2018icuas}, the designed controller assigns three quadrotors as leaders and the remaining robots as followers in the team. 
When an external force is applied to the payload, a fixed step is given to the leaders' positions along the estimated force direction. 
In \cite{romano2022jgcd}, five quadrotors collaborate with a human operator to transport a point mass payload. 
The force applied by a human on the payload is estimated by summing the cable tension forces and subtracting the gravity. 
The human-applied force is fed into an admittance controller, which updates the desired quadrotor position and velocity in the formation. 
However, compared to our work, the cable tension magnitude is measured by a custom tension measurement module, and its direction by a motion capture system. 
Moreover, the aforementioned works present substantial limitations as they presume the payload to be a point mass and constrain the human operator to manipulate the payload in only 2D. 
On the contrary, in our proposed work, we widen the scope of physical interactions between the human operator and the payload to all $6$ DoF. 
This enhancement is achieved by modeling the payload as a rigid body with $6$ DoF and developing an estimator to estimate the full $6$ DoF wrench acting on the payload. 
Furthermore, our system removes the use of any force-measuring devices on the robots or payload (except to obtain the ground truth during testing for validating our estimation approach). 
These are unique characteristics that increase the flexibility and applicability of our solution compared to existing ones.

\begin{figure}[t]
\centering
  \includegraphics[width=\columnwidth]{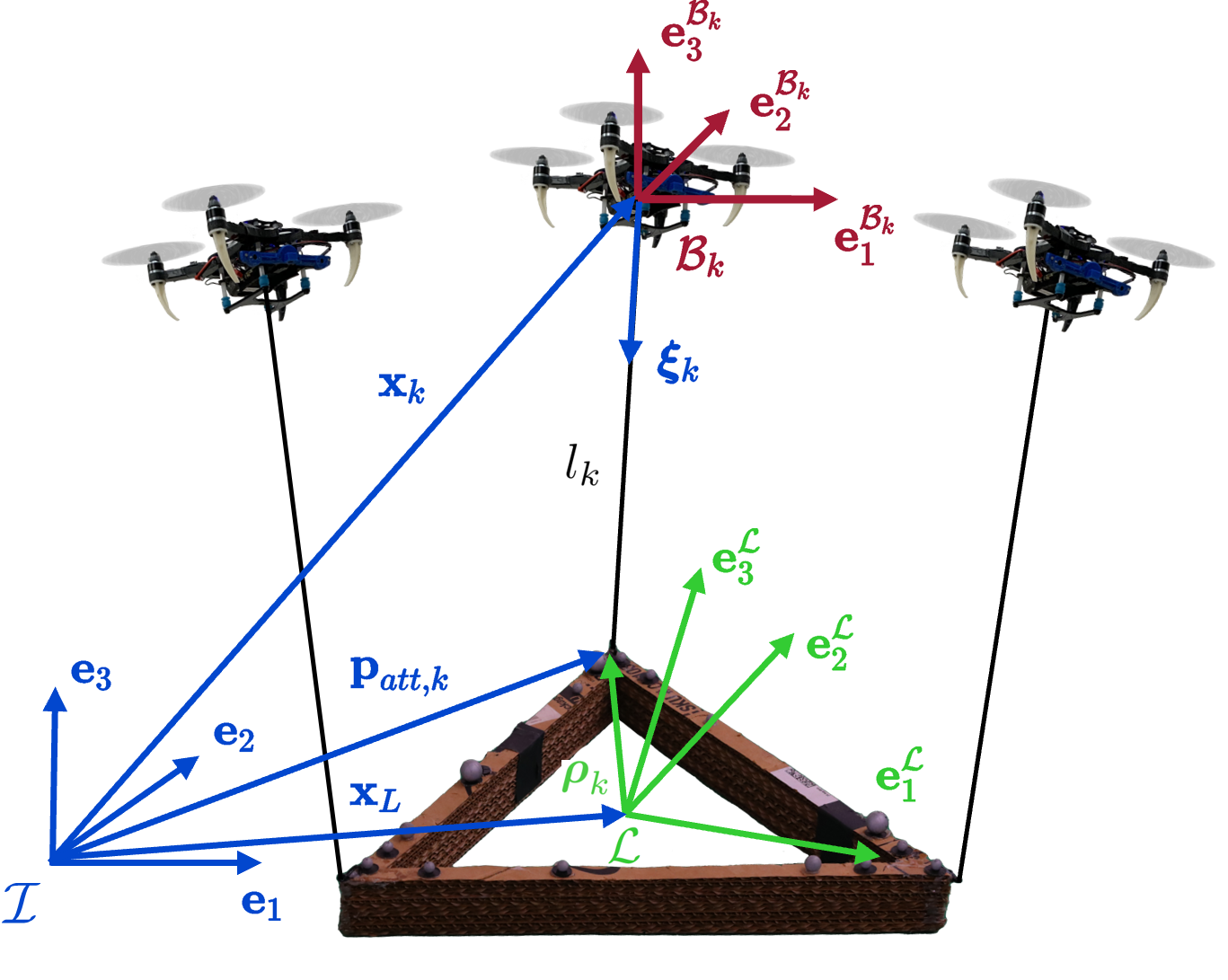}  
  \caption{System convention definition: $\worldf$, $\loadf$, $\robotf_{\id}$ denote the world frame, the payload body frame, and the $\id^{th}$ robot body frames, respectively, for a generic quadrotor team that's cooperatively transporting and manipulating a cable-suspended payload.\label{fig:frame_definition}}
\end{figure}

\begin{figure*}[t]
    \centering
    \includegraphics[width=\textwidth]{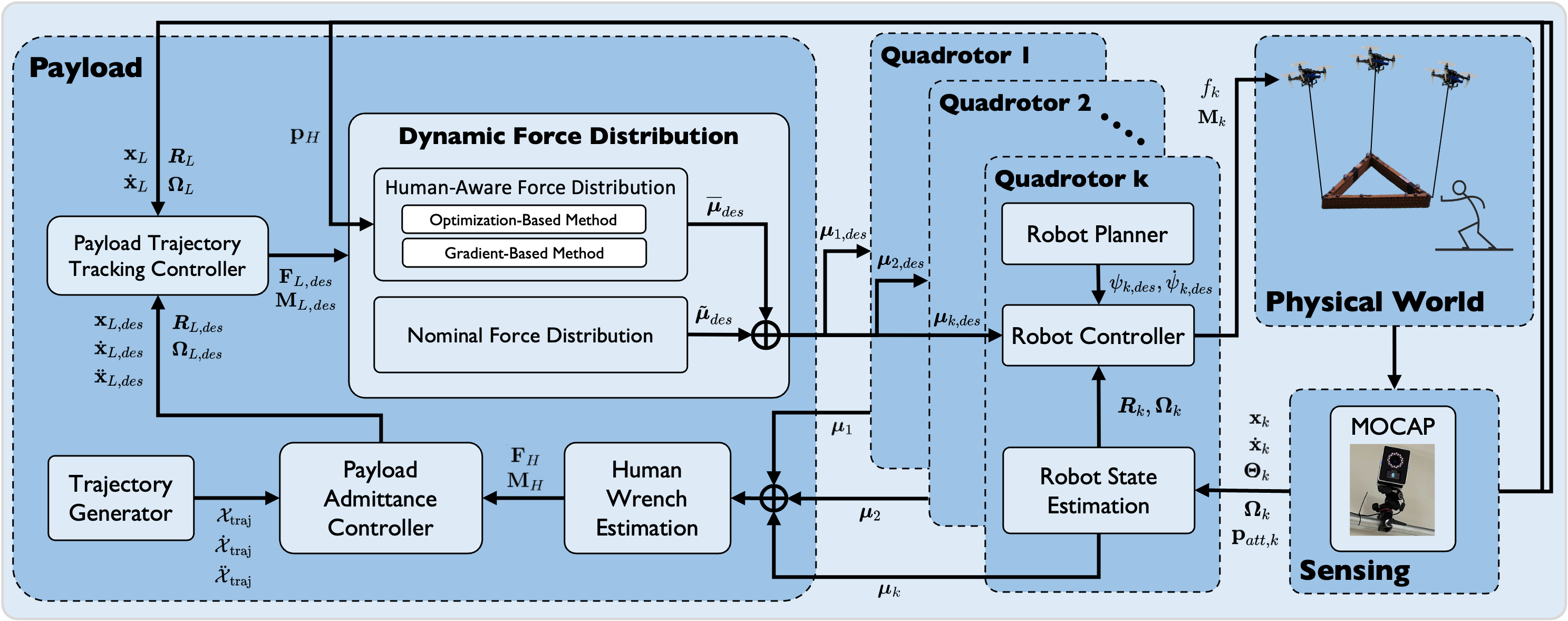}
    \caption{Block diagram of the system illustrating the overview of the system. It begins with a trajectory generator, which outputs the desired trajectory of the payload in all six degrees of freedom. The payload admittance controller updates this trajectory based on the estimated human input, adapting to the interaction wrench exerted by the human on the payload. The modified trajectory is then passed to the payload controller as the desired payload state to track. The payload controller calculates the desired wrench to control the payload's motion. Subsequently, the dynamic force distribution module allocates the desired cable forces based on the human's position and the system dynamics, and distributes the desired cable forces for each robot to track. Each robot controller will then track its corresponding desired cable force and output the corresponding thrust and moment commands to the quadrotor platform.}
    \label{fig:control_block_diagram}
\end{figure*}
\section{Overview}\label{sec:overview}
In this section, we present an overview of our proposed system designed to enable a team of quadrotors to collaborate with a human operator to manipulate a rigid-body payload. Depicted in Fig.~\ref{fig:frame_definition}, the system comprises $n$ quadrotors, and each quadrotor is tethered by a cable to its center of mass. The software architecture of the system encompasses couple of primary components: planning and control, and physical human-robot interaction, as shown in Fig.~\ref{fig:control_block_diagram}. We provide a more detailed description of each module below.

\subsection{Planning and Control}

The planning module generates a desired trajectory for the payload, encompassing both position and orientation. The trajectory is represented by a polynomial trajectory of time. By differentiating the polynomial with respect to time, we can obtain payload state's respective first and second derivatives. These derivatives represent the twist and acceleration of the payload pose. We use the similar planning method as the one we used in our previous work to generate the trajectory for payload~\cite{guanrui2022rotortm}. The desired values will be fed into the admittance controller. If human interaction with the payload occurs, the admittance controller will update the trajectory in response to the human inputs.

The control proposed in this paper, described in Section~\ref{sec:Control}, adopts a hierarchical design that comprises the payload tracking controller, dynamic force distribution, and robot controller. The hierarchical controller design offers several advantages. Firstly, it simplifies the complex task of controlling the entire system by breaking it down into several manageable subtasks. At the base of our design is the robot controller, which computes the control actions based on local information, such as cable direction and robot orientation. This strategy allows us to first test the individual robot controller with a single robot, thereby ensuring we do not risk compromising the entire system. After successfully building and testing the robot controller, we can proceed to test the higher-level modules, including the payload controller and dynamic force distribution. 

Moreover, the modularity enhances the system's maintainability and scalability, as changes or improvements to one part of the control system can be implemented without affecting other components. For example, in our case, we can introduce the adaptation module at the payload level without changing the robot controller.

In the following sections, we will provide a more detailed overview of the controllers within the control design.

\subsubsection{Payload Controller}
The payload controller's function is to control the payload's state, enabling it to track the adapted desired payload states from the admittance controller. It generates the desired manipulation force and moment on the payload to track the desired payload states.
\subsubsection{Dynamic Force Distribution}
The dynamic force distribution dynamically distributes the desired cable tension forces that each quadrotor needs to exert to manipulate the payload. It consists of two parts: nominal force distribution and human-aware force modification.

Nominal Force Distribution:
The nominal module processes the desired forces and moments on the payload, as computed by the payload controller. It maps the desired wrench on the payload into nominal desired cable tension forces for each quadrotor.

Human-Aware Force Distribution:
This module leverages the redundancy inherent in a multi-quadrotor system to adjust the nominal cable tension forces calculated by the nominal tension distribution module. It modifies the nominal cable tension forces without impacting the overall manipulation forces and moments exerted on the payload computed by the payload controller.
\subsubsection{Robot Controller}
Each quadrotor runs an individual robot controller to track the desired cable tension forces assigned from the dynamic force distribution module. It commands thrust and moment to the quadrotor such that the quadrotor can track the desired cable direction as well as exert the desired tension in the cable.
\subsection{Physical Human-Robot Interaction}
The Human Interaction module proposed in this paper in Section~\ref{sec:admittance}, comprises two main parts: the admittance controller and Human Input Estimation.

\subsection{Robot State Estimation}
Each robot runs an onboard Unscented Kalman Filter(UKF) to estimate the robot's position, velocity, orientation, angular velocity, and the cable force applied to the robot. The cable force estimated by each quadrotor will be shared among the team to collaboratively estimate the human wrench. 

\subsubsection{Human Wrench Estimation}
This component estimates the human's input wrench on the payload by collecting all the cable forces estimated by each quadrotor. Given the challenges associated with adding additional force sensors to the payload for real-world applications, we aim to minimize modifications to maintain cost-effectiveness. Our innovative solution, which does not require any additional force sensor, involves each quadrotor estimating its cable force. And the estimated cable force is then shared among the team to deduce the total forces and moments from all cables, further derive the human input wrench.

\subsubsection{Payload Admittance Controller}
The admittance controller modifies the desired payload trajectory based on human-applied forces and moments. This would allow the system to adapt to the human input, leading to collaborative transportation and manipulation of the payload between the human and the team of quadrotor robots. 

\begin{table}[t]
\caption {Notation table\label{tab:notation}} 
\centering
%\newcolumntype{s}{}
\begin{tabularx}{0.48\textwidth}{>{\hsize=0.58\hsize}X >{\hsize=1.42\hsize}X}
    \hline\hline
 $\worldf$, $\loadf$, $\robotf_{\id}$ & world frame, payload frame, $\id^{th}$ robot frame\\
 $\loadmass,\massi      \in\realnum{}$ &  mass of payload, $\id^{th}$ robot\\
 $\loadpos,\robotpos{\id}\in\realnum{3}$ &  position of payload, $\id^{th}$ robot in $\worldf$\\
 $\loadvel, \loadacc    \in\realnum{3}$ & linear velocity, acceleration of payload in $\worldf$\\
$\robotvel{\id}, \robotacc{\id}\in\realnum{3}$ &linear velocity, acceleration of $\id^{th}$ robot in $\worldf$\\
 $\loadrot              \in\SOt$&orientation of payload with respect to $\worldf$ \\
 $\robotrot{\id}        \in\SOt$&orientation of $\id^{th}$ robot with respect to $\worldf$ \\
 $\robotypr{\id}        \in\realnum{3}$&vector of $\id^{th}$ robot's yaw, pitch, roll in $\worldf$ \\
 $\loadangvel$, $\loadangacc\in\realnum{3}$ & payload's angular velocity, acceleration in $\loadf$\\
  $\robotangvel{\id}$, $\robotangacc{\id}\in\realnum{3}$& $\id^{th}$ robot's angular velocity, acceleration in $\robotf_{\id}$\\
 $\inertiaload,\inertia_{\id}\in\realnum{3\times3}$   &  moment of inertia of payload, $\id^{th}$ robot\\%\hline
 $\cableveci            \in S^2$&unit vector from $\id^{th}$ robot to attach point in $\worldf$\\
 $\omegai               \in\realnum{3}$, $l_{\id}\in\realnum{}$&angular velocity, length of $\id^{th}$ cable\\
 $\scalartension{\id}   \in\realnum{}$ & tension magnitude within the $\id^{th}$ cable\\
 $\extplF,\netplforce   \in\realnum{3}$ & external human force, net force on payload in $\worldf$\\
 $\extplM               \in\realnum{3}$ & external human moment on payload in $\loadf$\\
 $\netplmoment          \in\realnum{3}$ & net moment on payload in $\loadf$\\
 $f_{\id}              \in\realnum{}$&total thrust of $\id^{th}$ quadrotor\\
 $\qdF{\id} \in\realnum{3} $  &control force on $\id^{th}$ robot in $\worldf$\\
 $\qdM{\id} \in\realnum{3} $  &control moment on $\id^{th}$ robot in $\robotf_{\id}$\\
 $\vecrho{\id}          \in\realnum{3} $ &$\id^{th}$ attach point position in $\loadf$\\
 $\Phuman, \patt{\id}  \in\realnum{3}$ & human position, $\id^{th}$ attach point position in $\worldf$\\
    \hline\hline
\end{tabularx}
\end{table}

\section{System Dynamics}\label{sec:Dynamics}
This section presents the modeling of the overall system dynamics. We consider a scenario where a team of $n$ quadrotors cooperatively manipulates a rigid body payload, as illustrated in Fig.~\ref{fig:frame_definition}. We establish the world frame $\worldf$ on the ground. The payload frame $\loadf$ is located at the payload's center of mass. The payload's position and orientation relative to $\worldf$ are denoted by $\loadpos$ and $\loadrot$, respectively.

The relevant variables in this paper are summarized in Table~\ref{tab:notation}. We denote the three elements of any 3D vector using subscripts $*_{x,y,z}$.

The system dynamics models are developed based on the following assumptions
\begin{enumerate}
    \item Aerodynamic interactions with the ground and other effects caused by high robot velocity are ignored due to its insignificant effect at a low moving speed that's achieved by this system;\label{assumption:aero}
    \item Each cable is assumed to be attached at the center of mass of each robot, each robot's center of gravity coincides with its geometrical center, and all cables are assumed to be massless with no dynamic effects on the system;\label{assumption:cable}
    \item Wind disturbances are ignored, the human operator would only interact with the payload, and all external forces on the payload and each robot are considered to be exerted by a human operator.\label{assumption:human-operator}
\end{enumerate}
Assumption~\ref{assumption:aero} is justified by the operational velocity, and spatial separation among agents we propose in Section~\ref{sec:nullspace_exploit}, which minimizes aerodynamic effects. Assumption \ref{assumption:cable} is based on the symmetrical design of the MAV and the lightweight nature of the cables. Finally, Assumption \ref{assumption:human-operator} is relevant due to our focus on indoor operation. 

\subsection{Basic Geometry}
As shown in Fig.~\ref{fig:frame_definition}, the $\id^{th}$ quadrotor attached one massless cable with length $l_k$ from its center of mass to the $\id^{th}$ attach point on the payload. The location of the attach point $\id$ with respect to $\loadf$ and $\worldf$ is represented by constant vector $\vecrho{\id}\in\realnum{3}$ and vector $\patt{\id}\in\realnum{3}$ respectively. Hence, from the geometry, we can have
\begin{equation}
    \patt{\id} = \loadpos +\loadrot\vecrho{\id}\label{eq:attach-point-position}.
\end{equation}
As we can also observe from Fig.~\ref{fig:frame_definition}, when the cable is taut, the robot's position can be obtained by using the attach point position and the cable as follows
\begin{equation}
    \robotpos{\id} = \patt{\id}-l_{\id}\cablevec{\id}\label{eq:attach-point-position},
\end{equation}
where $\robotpos{\id}$ is the $\id^{th}$ quadrotor's position in $\worldf$, and $\cablevec{\id}$ is the unit vector that represents the cable direction from the robot to the corresponding attach point in $\worldf$.

\subsection{Payload Dynamics}
The net force $\netplforce$ in $\worldf$ and moment $\netplmoment$ in $\loadf$ on the payload is determined by all the cable tension forces $\tension{\id}$, $\id=1, \cdots, n$, gravitational pull $\vecg$, and external wrench $\vecF_H, \vecM_H$ applied by the human operator
\begin{equation}
\begin{bmatrix}\netplforce\\\netplmoment\end{bmatrix}  =\begin{bmatrix}\extplF\\\extplM\end{bmatrix}+\matP\tension{}-\begin{bmatrix}
    \loadmass\vecg\\\mathbf{0}
    \end{bmatrix},\hspace{1em} \tension{} = \begin{bmatrix}\tension{1}\\\vdots\\\tension{n}\end{bmatrix},
    \label{eq:total_wrench_on_payload}
\end{equation}
where $\loadmass$ is the payload mass, $\vecg = g\axis{3}{}$, $g = 9.81 \si{m/s^2}$, $\axis{3}{} = \begin{bmatrix}0&0&1\end{bmatrix}^{\top}$.
In eq.~\eqref{eq:total_wrench_on_payload}, the matrix $\matP\in\realnum{6\times3n}$ maps tension vectors of all $n$ MAVs in $\worldf$ to the wrench on the payload with force in $\worldf$ and moments in $\loadf$
\begin{equation}
\begin{split}
{\matP} =\begin{bmatrix}\matI_{3\times3} & \matI_{3\times3} &\cdots &\matI_{3\times3}\\                                \hatvecrho{1}\loadrot^{\top}&\hatvecrho{2}\loadrot^{\top}&\cdots&\hatvecrho{n}\loadrot^{\top}\end{bmatrix}.
\end{split}
\label{eq:P_mat}
\end{equation}
where $\matI_{3\times3}\in\realnum{3\times3}$ is an identity matrix and the hat map $\hat{\cdot} : \realnum{3} \rightarrow \mathfrak{so}(3)$ is defined such that $\hat{\veca}\vecb = \veca \times \vecb, \forall \veca, \vecb \in \realnum{3}$.

By inspecting the matrix $\matP$, we observe that $\matP$ has 6 rows, independent of the number of robots in the system. For $n \geq 3$ robots, the number of columns of $\matP$ surpasses the number of rows, causing the dimension of the domain of $\matP$ to exceed that of its image. Hence, there is an additional nullity in matrix $\matP$, which can be represented by the null space of the matrix $\matP$, denoted as $\mathcal{N}(\matP)\subset\realnum{3n-6}$. The system can utilize the nullity to accomplish secondary tasks, such as obstacle avoidance, inter-robot separation, or keeping a distance between human and robot~\cite{guanrui2023iros}, which we will introduce in Section~\ref{sec:Control}. 

%With any number of robots $n \geq 3$, the system allows us to exploit the system's additional redundancy, quantified as $\realnum{3n-6}$. effectively represents this redundancy. We will utilize this property in Section~\ref{sec:Control} to maintain minimum distance among the human operator and the robots during the manipulation task.

% $\loadrot\in\realnum{3\times3}$ is the payload's rotation matrix with respect to the frame $\worldf$, and $\vecrho{\id}\in\realnum{3}$ is the distance vector from the payload's center of mass to the $\id^{th}$ attach point with respect to the payload frame $\loadf$.

Through standard rigid body dynamics, we can obtain the translational and rotational dynamics of the payload as
\begin{equation}
    \loadmass\loadacc{} = \mathbf{F}_L,\hspace{1em}
    \inertiaload\loadangacc = \mathbf{M}_L-\loadangvel\times\inertiaload\loadangvel,
    \label{eq:payload_dynamics}
\end{equation}
where $\loadacc, \loadangacc\in\realnum{3}$ is the payload linear and angular acceleration respectively, $\loadangvel\in\realnum{3}$ is the payload angular velocity and $\inertiaload\in\realnum{3\times3}$ is the payload's inertia matrix. 

\subsection{Quadrotor Dynamics}\label{sec::quadrotor-dynamics}
Based on assumptions 2 and 3, we consider the translational and rotational dynamics of the $\id^{th}$ quadrotor as follows
\begin{equation}
\massi\robotacc{\id} =\inputforce{\id}-\tension{\id}-\massi\vecg, \label{eq:quad_eom_trans_dropped}
\end{equation}
\begin{equation}
\inertia_{\id}\robotangacc{\id} = \qdM{\id}-\robotangvel{\id}\times\inertia_{\id}\robotangvel{\id},\label{eq:quad_eom_rot_dropped}
\end{equation}
where $\inputforce{\id}$ and $\qdM{\id}$ are the control force and moment on the $\id^{th}$ quadrotor, $\tension{\id}=-\scalartension{\id}\cableveci$ is the tension force applied on the $\id^{th}$ quadrotor. 

When the cable is taut, the motion of $\id^{th}$ quadrotor is constrained to the surface of a sphere centered at the $\id^{th}$ attach point, with a radius equal to the cable length~\cite{Leecdc2014}. The dynamics of the cable direction can then be derived using the Lagrange d'Alembert principle as follows~\cite{guanrui2021iser}: 
\begin{equation}
\cableddotvec{\id} = \frac{1}{\massi l_{\id}} \hat{\cablevec{}}_{\id}^2\prths{\inputforce{\id}-\massi\acci} - \twonorm{\cabledotveci}^2\cableveci ,\label{eq:cable_dynamics}
\end{equation} 
where $\veca_{\id}$ is the acceleration of the $\id^{th}$ attachment point
 \begin{equation}
 \veca_{\id}=\loadacc + \vecg-\loadrot\hatvecrho{\id}\loadangacc+\loadrot\hat{\angvel}_L^{2}\vecrho{\id}.
\end{equation}
\section{Control}\label{sec:Control}
In this section, we introduce a hierarchical nonlinear controller that enables a team of $n$ quadrotors to manipulate a rigid-body load suspended by cables. The formulation of the controller is based on the system dynamics presented in Section~\ref{sec:Dynamics}. Fig.~\ref{fig:control_block_diagram} illustrates the hierarchical structure of the controller.

The hierarchy begins with a payload controller, detailed in Section~\ref{sec:payload_controller}, which generates the desired wrenches $\plFd,\plMd$ to control the position and orientation of the payload. Subsequently, the dynamic force distribution module, described in Section~\ref{sec:force_allocation}, assigns desired cable force vectors $\tension{\id}, \id=1,\cdots,n$ for each robot, based on the desired payload wrench $\plFd, \plMd$. The control hierarchy finishes with the robot controller at its lowest level, where the individual robot controller on each robot tracks its corresponding desired cable force vector $\tension{\id}$. Each robot controller, associated with the $\id^{th}$ robot, computes the appropriate thrust and moment commands for the robot, as further shown in Section~\ref{sec:robot_controller}.

\subsection{Payload Controller}\label{sec:payload_controller}
We present a payload controller that enables the load to follow the desired trajectory in a closed loop. The subscript $*_{des}$ denotes the desired value given by the trajectory planner. The desired forces and moments acting on the payload are designed as  
\begin{equation}
\label{eq:payloadtotalforce}
\vecF_{L,\mathit{des}} = \loadmass\veca_{L,c},
\end{equation}
\begin{equation*}
\veca_{L,c} =  \matK_{p}\vece_{\loadpos}+\matK_{d}\vece_{\loadvel}+\matK_{i}\int_{0}^{t}\hspace{-5pt}\vece_{\loadpos}d\tau + \loadaccdes + \vecg,
\end{equation*}
\begin{equation}
\begin{split}
\mathbf{M}_{L,\mathit{des}} =& \matK_{\loadrot}\vece_{\loadrot}+\matK_{\loadangvel}\vece_{\loadangvel} +\inertiaload\loadrot^{\top}\loadrotdes\loadangaccdes \\&+\prths{\loadrot^{\top}\loadrotdes\loadangveldes}^{\wedge}\inertiaload\loadrot^{\top}\loadrotdes\loadangveldes, \label{eq:payloadtotalmoment} 
\end{split}
\end{equation}
where $\matK_{p},\matK_{d},\matK_{i}$, $\matK_{\loadrot},\matK_{\loadangvel}\in\realnum{3\times3}$ are constant diagonal positive definite matrices, and 
\begin{equation}
\begin{split}
\vece_{\loadpos} &= \loadposdes - \loadpos,~\vece_{\loadvel} = \loadveldes - \loadvel, \\
\vece_{\loadrot} &= \half\prths{\loadrot^{\top}\loadrotdes-\loadrotdes^{\top}\loadrot}^{\vee}, \\
\vece_{\loadangvel} &=  \loadrot^{\top}\loadrotdes\loadangveldes - \loadangvel.\label{eq:control_errors}
\end{split}
\end{equation}
In the above equation, the vee map $^\vee:\sot\rightarrow\realnum{3}$ is the reverse of the hat map $\hat{\cdot}$.

\subsection{Dynamic Force Distribution}~\label{sec:force_allocation}
In this section, we will present our dynamic force distribution method that allocates the desired force $\plFd$ and moment $\plMd$ on the payload to the desired cable tension forces $\tension{\id}, \id=1,\cdots,n$. The force distribution comprises two segments: the nominal force distribution and the human-aware force distribution. 

The nominal force distribution, originated from nonlinear geometric control method~\cite{guanrui2021ral,guanrui2021iser}, distributes the desired payload wrench to the desired cable tension forces using minimum norm solution. 

On the other hand, the human-aware force distribution leverages the system redundancy, given that the robot number $n\geq3$, to obtain the cable forces that yield a zero effective wrench on the payload. This adjustment enables the system to modify the desired cable forces from the nominal distribution for secondary objectives, such as maintaining a distance from the robots to the human operator and between the robots themselves, without impacting the original manipulation tasks.

In the following, we will first present the nominal force distribution method in Section~\ref{sec:nominal_tension_distribution}, followed by the human-aware force distribution method in Section~\ref{sec:nullspace_exploit}.

\subsubsection{Nominal Force Distribution}\label{sec:nominal_tension_distribution}
Once the desired payload wrench $\vecF_{L,des}$, $\matM_{L,des}$ is obtained, it can be distributed to the desired tension force $\bar{\vecmu{}}_{\id,des}$ along each cable as 
\begin{equation}
  \bar{\vecmu{}}_{des} = \begin{bmatrix}\bar{\vecmu{}}_{1,des}\\
            \vdots\\
            \bar{\vecmu{}}_{n,des}\end{bmatrix} = \matP^{\dagger}\begin{bmatrix}\vecF_{L,des}\\
             \mathbf{M}_{L,des}\end{bmatrix},
             %\matP^{\top}\prths{\matP\matP^{\top}}^{-1}
    \label{eq:destension}
\end{equation}
where ${\matP}^{\dagger} = {\matP}^{\top}({\matP}{\matP}^{\top})^{-1}$ is the Moore-Penrose inverse of $\matP$. The above solution can be directly used as the desired cable tension vector for the robot, like in our previous works~\cite{guanrui2021iser,guanrui2021ral}. However, the above solution does not exploit the possibility of the quadrotor
team’s needs to accomplish secondary tasks. For example, the second task can be avoiding obstacles or, as shown in this paper, spatially separating the human and the robots during physical collaboration. 

\subsubsection{Human-Aware Force Distribution}\label{sec:nullspace_exploit}
As we have mentioned before, the human-aware force distribution exploits the system redundancy to modify the desired cable forces from the nominal distribution for secondary tasks. The secondary tasks can be maintaining a distance from the robots to the human operator and between the robots themselves, without impacting the original manipulation tasks.

To accomplish these, we propose and discuss two distinct approaches in this section. The system aims to allocate the wrench to the desired cable tension forces with two principal goals in mind:
\begin{enumerate}
    \item Minimize the total cable tension forces to conserve the robot's energy.
    \item Utilize the null space of the cable distribution matrix $\matP$ to facilitate secondary tasks, such as ensuring a safety distance between the robots and potential human operators within the team.
\end{enumerate}

In the following, we introduce two methods enabling each MAV to maintain distance from an object. Concurrently, these methods allow the team to maintain the original desired load forces, $\vecF_{L,\mathit{des}}$, and moments, $\mathbf{M}_{L,\mathit{des}}$, as outlined in Section~\ref{sec:payload_controller}. This approach ensures that the original objectives in load manipulation are not compromised.

The methods leverage the redundancy inherent in system configurations involving more than three MAVs, as explained in Section~\ref{sec:Dynamics}. We intend to find a desired tension force modifier, $\nullspacev{des}\in\mathcal{N}({\matP})$, that modifies $\bar{\vecmu{}}_{des}$ in eq.~(\ref{eq:destension}), and satisfies
\begin{equation}
    {\matP}\nullspacev{des}=\bold{0},\hspace{1em}\nullspacev{des} = \begin{bmatrix}\nullspacev{1,des}\\\vdots\\\nullspacev{n,des}\end{bmatrix}.\label{eq:nullspace_cond}
\end{equation}
Eq.~\eqref{eq:nullspace_cond} means the tension modifiers result in zero wrench on the payload, which doesn't affect the original desired manipulation wrench from eqs.~\eqref{eq:payloadtotalforce} and \eqref{eq:payloadtotalmoment}. With the tension force modifier, we can update $\id^{th}$ robot's desired cable tension force as
\begin{equation}
    \tension{\id,des} = \bar{\vecmu{}}_{\id,des}+\nullspacev{\id,des}.
    \label{eq:nullspace_equation}
\end{equation}
Intuitively, $\mathcal{N}({\matP})$ provides all the possible combinations of $n$ cable tension vectors that can generate internal motions of the structure (i.e., variations of the cables' directions) that do not affect the load configuration controlled by the method presented in Section~\ref{sec:payload_controller}. This is confirmed by eq.~(\ref{eq:total_wrench_on_payload}) and eq.~(\ref{eq:nullspace_equation}), as $\bar{\vecmu{}}_{\id,des}$ would create a nonzero net wrench on the payload while $\nullspacev{\id,des}$ creates zero net wrench. Moreover, $\nullspacev{\id,des}$ can be related to the position of each robot
\begin{equation}
    \patt{\id}+l_{\id}\cablevecnull{\id,des}=\robotpos{\id},
    \label{eq:nullspace_to_position}
\end{equation}
and 
\begin{equation}
\cablevecnull{\id,des}=-\cablevec{\id,des} =\frac{\bar{\vecmu{}}_{\id,des}+\nullspacev{\id, des}}{\|\bar{\vecmu{}}_{\id,des}+\nullspacev{\id, des}\|}.
\end{equation}
The advantage is that we can exploit $\nullspacev{\id,des}$ to enforce the $\id^{th}$ robot to maintain a certain distance with respect to the other agents in the system and other objects in the environments, like a potential human operator.

Therefore, the human-aware force distribution needs to find the aforementioned tension force modifier,  $\nullspacev{des} \in \mathcal{N}({\matP})$, and use eq.~(\ref{eq:nullspace_equation}) to move MAVs based on eq.~(\ref{eq:nullspace_to_position}) without affecting the payload. We propose the following two approaches for finding $\nullspacev{des}$
\begin{enumerate}
    \item \textit{Gradient-Based Method}: Find a $\nullspacev{des}$ such that each MAV maximizes the distance between itself and the object using a gradient ascent method.
    \item \textit{Optimization-Based Method}: Find a $\nullspacev{des}$ such that each MAV guarantees a predetermined minimal safe distance between all its neighboring drones and the object by using nonlinear optimization. 
\end{enumerate}

In the following, we can describe the obstacle or human operator as a particular object of interest in the environment. The corresponding point position with respect to $\worldf$ is denoted as $\mathbf{p}_O$ in the following controller formulation. 

\textbf{Gradient-Based Method:} 
Inspired by strategies used for redundant rigid link robot arms in \cite{siciliano_robotics_2009}, we introduce a gradient-based method to compute $\nullspacev{des}$. Specifically, a pseudo tension force modifier, $\pseudotensionm{des}$, is found by maximizing the distance between objects in the environment and each drone. $\pseudotensionm{des}$ is then projected into $\mathcal{N}({\matP})$ to become the tension force modifier $\nullspacev{des}$. To update $\pseudotensionm{des}$, we propose 
\begin{equation}
    \pseudotensionm{des}=\softlimit\frac{\partial \bold{w}(\nullspacev{des})}{\partial\nullspacev{des}},
    \label{eq:findg}
\end{equation}
where $\softlimit\in\realnum{3n\times 3n}$ is a diagonal positive-definite matrix with variable and tunable coefficients on its diagonal, and $\bold{w}(\nullspacev{des})$ is the cost function. $\bold{w}(\nullspacev{des})$ is defined as the squared distance between each drone and the object from which the system is required to keep a safe distance as follow
\begin{equation}
    \bold{w}(\nullspacev{des})=\sum_{i=1}^{k}\left\|\Pobject- \left(\patt{\id}+\bold{l}_{\id}\cablevecnull{\id,des}\right)\right\|^2,
    \label{eq:costfunction}
\end{equation}
where $\mathbf{l}_{\id}=l_{\id}\matI_{3\times3}$. Note that $\cablevecnull{\id,des}$ points from the $\id^{th}$ attach point to the $\id^{th}$ robot.
We can now compute the partial derivative of the cost function corresponding to the $\id^{th}$ robot by using eq.~(\ref{eq:costfunction}) and obtain the following:
\begin{equation}
\begin{split}
\frac{\partial \bold{w}(\nullspacev{\id,des})}{\partial \nullspacev{\id,des}}=&-2\bold{l}_{\id}\left[\Pobject-\patt{\id}\right]^{\top}\frac{\partial\cablevecnull{des}}{\partial\nullspacev{\id,des}}\\
% Second line of the equation
=& -2\bold{l}_{\id}\left[\Pobject-\patt{\id}\right]^{\top}\frac{(\matI_{3\times3}-\cablevecnull{\id,des}\cablevecnull{\id,des}^\top)}{\|\bar{\vecmu{}}_{\id,des}+\nullspacev{\id,des}\|}.
                            % =& -2\bold{l}_{\id}\left[\Phuman-\patt{\id}\right]^{\top}\\&\left[\frac{1}{\|\tension{\id}+\pseudotensionm{\id}\|}\matI_{3n\times3n}-\frac{(\tension{\id}+\pseudotensionm{\id})(\tension{\id}+\pseudotensionm{\id})^{\top}}{\|\tension{\id}+\pseudotensionm{\id}\|^3}\right]
\end{split}
\label{eq:robotkgrad}
\end{equation}
For each control step, we update $\pseudotensionm{des}$ based on eq.~(\ref{eq:findg}), performing gradient ascent to maximize the distance between each robot and the objects in the environment.
To regulate the effect of gradient on each robot when the object is far away, we propose each element of $\softlimit$ as an exponential decay function of the robot-to-object distance
\begin{equation}
    \softlimit=\text{diag}\prths{\mathbf{Q}_1, \hdots ,\mathbf{Q}_n},
    \hspace{1em}\mathbf{Q}_{\id}= ae^{-b\|\Pobject-\robotpos{\id}\|}\matI_{3\times3},
\end{equation}
where, $a, b\in\realnum{}$ are tunable coefficients. With these varying coefficients, we can implicitly impose distance limits as the gradients will only have impact on the tension modification when $\id^{th}$ robot is close enough to the object. However, till now, $\pseudotensionm{des}$ from eq.~(\ref{eq:findg}) is not yet in $\mathcal{N}({\matP})$. Hence we consider the following optimization to project $\pseudotensionm{des}$ into the null space of the matrix $\matP$
\begin{equation}
\begin{aligned}
    \min_{\bar{\tension{des}}} \quad & \|\pseudotensionm{des}-\nullspacev{des}\|^2 \\
    \text{s.t.}      \quad & {\matP}\nullspacev{des}=0.\label{eq:grad_method_opt_formulation}
    \end{aligned}
\end{equation}
Since the above optimization problem is a quadratic programming problem with linear equality constraints, there is a closed-form solution shown as follows\cite{Bertsekas/99}  
\begin{equation}
    \nullspacev{des} = \nullproj\pseudotensionm{des}= \prths{\boldsymbol{I}-{\matP}^{\dagger}{\matP}}\pseudotensionm{des},
    \label{eq:nullprojector}
\end{equation}
where ${\matP}^{\dagger}$ is the pseudoinverse as in eq.~(\ref{eq:destension}), and $\nullproj$ is an orthogonal projector that projects any $\pseudotensionm{des}$ orthogonally into the null space of ${\matP}$. 

Using this result, we project $\pseudotensionm{}$ into $\mathcal{N}({\matP})$ with eq.~(\ref{eq:nullprojector}), ensuring zero additional wrenches being applied on the payload when each robot is maximizing distance from the object. Finally, we update the desired tension vector as
\begin{equation}
    \tensiondesall =  \bar{\vecmu{}}_{des} + \nullspacev{des} = \bar{\vecmu{}}_{des} + \nullproj\pseudotensionm{des}. \label{eq:gradient_based_safety_controller_final_result}
\end{equation}
%The $\pseudotensionm{}$ therefore locally maximizes the distance from the object compatibly with the primary task defined in eq.~(\ref{eq:destension}).

\textbf{Optimization-Based Method:}
In this section, we directly formulate an optimization problem to solve for a tension force modifier $\nullspacev{des}$ in $\mathcal{N}({\matP})$ that guarantees safety distance among the objects and the robots. The nonlinear optimization problem is to minimize the total square norm of the resulting cable tension vector. Furthermore, we formulate $n$ robot-to-object distance constraints, as well as another ${n \choose 2}=\frac{n(n-1)}{2}$ constraints are added to prevent each pair of robots from collision.
Consider the following nonlinear optimization problem
\begin{equation}
\begin{aligned}
    \min_{\vecc}         \quad      & \left\|\bar{\vecmu{}}_{des}+\matN\veclamb\right\|^2 \\
    \textrm{s.t.}    \quad      & \|\Pobject-\robotpos{k}\|^2 \geq \humanlimit^2,  && 0<k\leq n,\\
                                & \left\|\robotpos{i}-\robotpos{j}\right\|^2 \geq \robotlimit^2,   && 0<i<j\leq n,\\
    \end{aligned}
    \label{eq:nlopt}
\end{equation}
where the columns of $\matN$ spans $\mathcal{N}({\matP})$ and $\vecc\in\realnum{3n-6}$ is the vector to be optimized. $\robotlimit$ and $\humanlimit$ are two scalar values denoting the predetermined safe minimum distance allowed between robots and between the object and each robot, respectively. The $\id^{th}$ robot's position is expressed in terms of $\matN_{\id}\veclamb$ and $\bar{\vecmu{}}_{\id,des}$
\begin{equation}
    \robotpos{\id} = \patt{\id}+l_{\id}\frac{\bar{\vecmu{}}_{\id,des}+\matN_{\id} \veclamb}{\|\bar{\vecmu{}}_{\id,des}+\matN_{\id}\veclamb\|},
\end{equation}
where $\matN_{\id}$ represents the three rows from the $\id^{th}$ row to the $\id+2^{th}$ row of the null space basis matrix $\matN$, which corresponds to the $\id^{th}$ MAV. Since eq.~(\ref{eq:nlopt}) is a nonlinear optimization problem with quadratic cost function and quadratic constraints, we use sequential quadratic programming solver for nonlinearly constrained gradient-based optimization~\cite{dieter1994acmnlsqp} in NLOPT~\cite{johnson2011nlopt} to solve eq.~(\ref{eq:nlopt}) and obtain $\vecc$. After obtaining $\vecc$, the desired cable tension forces can be obtained as follows
\begin{equation}
    \tension{des} = \bar{\vecmu{}}_{des} + \nullspacev{des} = \bar{\vecmu{}}_{des} + \matN \veclamb. \label{eq:opt_based_safety_controller_final_result}
\end{equation}

\textbf{Discussion:} The proposed methods are both effective for the quadrotor team to keep a safe distance away from a given object, as we also experimentally verify in Section~\ref{sec:experimental_results}. However, considering computational aspects, the gradient-based method requires fewer resources compared to the optimization-based method. This is primarily due to the closed-form solution offered by the gradient-based approach, as demonstrated by eq. \eqref{eq:robotkgrad}. On the other hand, the optimization-based method needs to solve a nonlinear optimization problem. In Section~\ref{sec:discussion}, we provide a quantitative analysis of the computational complexity and resource usage for both methods based on our implementation.

\subsection{Robot Controller}\label{sec:robot_controller}
In this section, we present the controller on each robot that enables the quadrotor to execute the desired cable tension force. The same robot controller has been used in our previous works~\cite{guanrui2021ral,guanrui2021iser}. 

Once we obtain the desired tension forces $\tension{des}$ from eq.~(\ref{eq:gradient_based_safety_controller_final_result}) or eq.~(\ref{eq:opt_based_safety_controller_final_result}), 
we can obtain the desired direction $\cablevecides$ and the desired angular velocity $\cableveldes{\id}$ of the $\id^{th}$ cable link as
\begin{equation*}
    \cablevecides = -\frac{\tensiondes{\id}}{\norm{\tensiondes{\id}}},\,\,\cableveldes{\id} = \cablevecides\times\cabledotvecides,
\end{equation*} 
where $\cabledotvecides$ is the derivative of the desired cable direction $\cablevecides$. After we obtain the desired cable direction $\cablevecides$ and cable angular velocity $\cableveldes{\id}$, we can determine the desired force vector for the robot $\inputforce{\id}$ as 
\begin{equation}
\begin{split}
\inputforce{\id} =& \inputpara{\id} + \inputperp{\id},\\
    \inputperp{\id}  =&\,\, \massi l_{\id}\cableveci\times\left[-\matK_{\cableveci}\vece_{\cableveci} -\matK_{\cablevel{\id}}\vece_{\cablevel{\id}} -\hat{\cablevec{}}_{\id}^2\cableveldes{\id}\right.\\
    &\left.-\prths{\cableveci\cdot\cableveldes{\id}}\cabledotvecides\right]- \massi\hat{\cablevec{}}_{\id}^2\veca_{\id,c},\\
\inputpara{\id}  =&\,\, \cableveci\cableveci^{\top}\tensiondes{\id} + \massi l_{\id}\twonorm{\cablevel{\id}}^2\cableveci  + \massi\cableveci\cableveci^{\top}\veca_{\id,c},\\
    \veca_{\mathit{\id},c} =&\,\,\veca_{L,c} -\loadrot\hatvecrho{\id}\loadangacc+ \loadrot\hat{\angvel}_{L}^2\vecrho{\id},
    \label{eq:control_inputs}
\end{split}
\end{equation}
where $\matK_{\cableveci}$ and $\matK_{\cablevel{\id}}\in\realnum{3\times3}$ are constant diagonal positive definite matrices, $\vece_{\cableveci}$ and $\vece_{\cablevel{\id}}\in\realnum{3}$ are the cable direction and cable angular velocity errors respectively
\begin{equation*}
\vece_{\cableveci}= \cablevecides \times\cableveci,~
\vece_{\cablevel{\id}} = \cablevel{\id} + \cableveci\times
\cableveci\times\cableveldes{\id}.
\end{equation*}
As we obtain the desired force vector of the quadrotor from~eq.~\eqref{eq:control_inputs}, we can follow~\cite{mellinger_snap_icra_2011} to derive the desired rotation $\robotrot{\id, des}$ and angular velocity $\bm{\Omega}_{\id,des}$ with desired yaw angle and desired yaw angular velocity from the robot's own planner. The 
thrust command $f_{\id}$ and moment command $\qdM{\id}$ to the $\id^{th}$ quadrotor are therefore selected as
\begin{align}\label{eq:ctrl_force_scalar}
    f_{\id} =&\,\, \inputforce{\id}\cdot\robotrot{\id}\axis{3}{} , \\\label{eq:ctrl_moment}
\qdM{\id} =&\,\, \mathbf{K}_{R}\mathbf{e}_{R_{\id}} + \mathbf{K}_{\bm{\Omega}}\mathbf{e}_{\bm{\Omega}_{\id}} + \bm{\Omega}_{\id}\times\inertia_{\id}\bm{\Omega}_{\id}\\
&-\inertia_{\id}\prths{\hat{\bm{\Omega}}_{\id}\robotrot{\id}^{\top}\robotrot{\id,des}\bm{\Omega}_{\id,des}-\robotrot{\id}^{\top}\robotrot{\id,des}\dot{\bm{\Omega}}_{\id,des}},\nonumber
\end{align} 
where $\matK_{R}$, $\matK_{\bm{\Omega}}\in\realnum{3\times3}$ are constant diagonal positive definite matrices, $\mathbf{e}_{R_{\id}}\in\realnum{3}$ and $\mathbf{e}_{\bm{\Omega}_{\id}}\in\realnum{3}$ are the orientation and angular velocity errors similarly defined using eq.~(\ref{eq:control_errors}). 
The readers can refer to~\cite{Leecdc2014} for stability analysis of the controller. % The aforementioned derivation addresses the challenges related to transitioning the nonlinear geometric controllers from theory to practice by showing a clear hierarchical formulation that can run on real robots. 
\section{Physical Human-Robot Interaction}\label{sec:admittance}
This section introduces the physical human-robot interaction module that enables a human operator to physically cooperate with a team of $n$ quadrotors in manipulating a suspended rigid-body payload. The module comprises two main sub-blocks: the estimation module and the admittance controller.

The estimation module is designed to facilitate the quadrotor team in estimating the human operator's input wrench exerted on the payload. The admittance controller takes the estimated human wrench and the desired payload state as input and generates a desired payload state to adapt the human's action. 

\subsection{Estimation}
We present the estimator design that allows the quadrotor team to estimate the external wrench applied to the payload by the human operator. First, in Section~\ref{sec:Robot_Estimator}, we introduce a quadrotor state estimator based on Unscented Kalman Filter (UKF) that runs onboard each quadrotor in a distributed fashion. Each quadrotor can leverage the estimator to estimate the cable force applied to it, without the need for a force sensor. Subsequently, in Section~\ref{sec:Wrench_Estimator}, we show how we can estimate the external wrench applied on the payload by the human operator via sharing the cable force on each quadrotor among the team. 

\subsubsection{Robot State Estimation}\label{sec:Robot_Estimator}
We consider the $\id^{th}$ MAV to have the following state $\state{\id}$
\begin{equation}
    \state{\id}=\begin{bmatrix}
    % \robotpos{\id} & \robotvel{\id} & \robotacc{\id} & \robotypr{\id} & \robotangvel{\id} & \robotangacc{\id}& \tension{\id} & \qdF{\id} & \qdM_{\id}
    \robotpos{\id}^{\top} & \robotvel{\id}^{\top}  & \robotypr{\id}^{\top} & \robotangvel{\id}^{\top}  & \cablevec{\id} & \cabledotvec{\id}^{\top} & \scalartension{\id}
    \end{bmatrix}^{\top},
\end{equation}
where $\robotypr{\id}\in\realnum{3}$ is a vector of the 3 Euler angles expressed according to the ZYX convention representing the robot's orientation. And the input is defined as
\begin{equation}
    \ekfinputi = \begin{bmatrix}f_{\id}&\qdM{\id}^{\top}\end{bmatrix}^{\top},
\end{equation}
where $f_{\id}$ and $\qdM{\id}$ are obtained based on motor speed measured by the electronic speed controllers on the robot. The relationship between motor speed and the resultant thrust and moment is expressed as follows:
\begin{equation}
\begin{bmatrix}
    f_{\id}\\\qdM{\id}
\end{bmatrix}=
\begin{bmatrix}
     k_f& k_f&k_f&k_f \\
     d_xk_f&    d_xk_f&  -d_xk_f& -d_x k_f\\
     -d_yk_f&  d_yk_f& d_yk_f& -d_yk_f\\
     k_m&         -k_m&      k_m&       -k_m\\ 
\end{bmatrix}
\begin{bmatrix}
    \omega_{m1}^2\\
    \omega_{m2}^2\\
    \omega_{m3}^2\\
    \omega_{m4}^2
\end{bmatrix}
    \label{eq:motor_speed}
\end{equation}
where $k_f$ and $k_m$ represent the motor constants corresponding to rotor force and moment, respectively. $d_x$ and $d_y$ denote the distances from the rotor to the body's x and y axes. Additionally, $\omega_{mj}$ signifies the motor speed of the $j^{th}$ motor. We denote the current time step as $*^t$ and the previous time step as $*^{t-1}$. Subsequently, we present the nonlinear process model and the linear measurement model of the Unscented Kalman Filter (UKF).
\paragraph{Process Model}
Based on MAV equations of motion presented in eq.~(\ref{eq:quad_eom_rot_dropped}), discretizations of quadrotor states are performed by assuming each control step moves forward in time by $\delta t$. The discrete-time nonlinear process model is
\begin{equation}
\begin{split}
\state{\id}^{t}\hspace{-2pt}=\hspace{-2pt}g(\state{\id}^{t-1}, \controlin{\id}^{t})=\hspace{-3pt}\begin{bmatrix}
   \robotpos{\id}^t \\ \robotvel{\id}^t  \\ \robotypr{\id}^t \\ \robotangvel{\id}^t  \\ \cablevec{\id}^t \\ \cabledotvec{\id}^t \\ \scalartension{\id}^t
    \end{bmatrix}
  \hspace{-3pt}=\hspace{-3pt}\begin{bmatrix}
    \robotpos{\id}^{t-1}+\robotvel{\id}^{t-1}\delta t+\robotacc{\id}^{t}\frac{\delta t^2}{2}\\
    \robotvel{\id}^{t-1}+\robotacc{\id}^{t}\delta t\\
    \left\lfloor\robotrot{\id}^{t-1}\textbf{exp}\left[\robotrot{\id}^{t-1}\robotangvel{\id}^{t-1}\delta t\right]\right\rfloor\\
    \robotangvel{\id}^{t-1}+\robotangacc{\id}^{t}\delta t\\
    \cablevec{\id}^{t-1}+\cabledotvec{\id}^{t}\delta t+\cableddotvec{\id}^{t}\frac{\delta t^2}{2}\\
    \cabledotvec{\id}^{t-1}+\cableddotvec{\id}^{t}\delta t\\
    \scalartension{\id}^{t-1}
    \end{bmatrix}.\hspace{-5pt}
\label{eq:nonlinear_process_model}
\end{split}
\end{equation}
For updating the Euler angles, a few nonlinear mappings are used as in \cite{sola2017quaternion}
\begin{enumerate}
    \item $\lfloor*\rfloor$ that maps $*\in\SOt$ to $\robotypr{}\in\realnum{3}$.
    \item $\textbf{exp}[*]$ that maps $*\in\realnum{3}$ to $\SOt$; or maps axis-angle $\mathfrak{so}(3)$, to rotation matrix $\SOt$. 
\end{enumerate} 
In eq.~(\ref{eq:nonlinear_process_model}), $\robotangvel{\id}^{t-1}$ is rotated into $\worldf$. After time step $\delta t$, the robot's angular displacement in $\worldf$, expressed in $\mathfrak{so}(3)$ is mapped into $\SOt$. This new robot orientation in $\SOt$ is then added to the previous orientation and the resultant orientation is converted to Euler angle using $\lfloor*\rfloor$. The equations of motion for unit cable direction are provided in \cite{guanrui2021ral}, with the results presented here 
\begin{equation}
\cableddotvec{\id}^t= 
\frac{\left(\hat{\cablevec{}}^{t-1}_{\id}\right)^2\prths{\inputforce{\id}-\massi\acci}}{\massi l_{\id}}-\twonorm{\cabledotveci^{t-1}}^2\cableveci^{t-1}\hspace{-2pt}.
\end{equation}
In our system, we employ a $16\times16$ time-invariant process noise diagonal covariance matrix, operating under the assumption that the system has zero-mean, additive Gaussian process noise. For the prediction phase, the Unscented Kalman Filter (UKF) algorithm utilizes eq.~(\ref{eq:nonlinear_process_model}). It's important to note that in instances where measurements are not available at every control time step, the UKF implements a nonlinear prediction method that assumes zero process noise. 

\paragraph{Measurement Model}
Using an indoor MOCAP system, we can measure everything in the state except tension magnitude. The measured states for $\id^{th}$ robot therefore are
\begin{equation}
    \measurement{\id}= \begin{bmatrix}
    % \robotpos{\id} & \robotvel{\id} & \robotacc{\id} & \robotypr{\id} & \robotangvel{\id} & \robotangacc{\id} & \qdF{\id} & \qdM_{\id}
    \robotpos{\id}^{\top} & \robotvel{\id}^{\top}  & \robotypr{\id}^{\top} & \robotangvel{\id}^{\top}  & \cablevec{\id}^{\top} & \cabledotvec{\id}^{\top}
    \end{bmatrix}^{\top}.
\end{equation}
We also want to note that using onboard Visual Inertial Odometry and vision-based methods from our previous work \cite{guanrui2021ral} can provide the same measurements as the motion capture does and can potentially make the entire measurement update process run fully onboard. However, this is out of the scope of this paper and we refer to it as future works. 

Denoting the time step when the UKF measurement update is triggered as $m$, the UKF finds the state prior to $m$, $\state{\id}^{m-1}$. A nonlinear propagation through eq.~(\ref{eq:nonlinear_process_model}) is performed by propagating sigma points around $\state{\id}^{m-1}$ through the system model in eq.~(\ref{eq:nonlinear_process_model}) as shown in~\cite{thrun_probabilistic_robotics}. 

The linear measurement model is
$$\measurementmodel =  \begin{bmatrix}
    \mathbf{I}_{18\times18}&\mathbf{0}_{18\times1}\\
    \mathbf{0}_{1\times18}&\mathbf{0}_{1\times1}\\
\end{bmatrix}.$$
The system state prediction is compared to the actual measurement using the above model
\begin{equation}
    \state{\id}^m =\kalmangain{\id}(\measurementmodel
    \statebar{\id}^{m}-\measurement{\id}),
\end{equation}
where $\statebar{\id}^{m}$ is the averaged state after sigma point propagation and $\kalmangain{\id}$ is the Kalman gain. Kalman gain is computed based on the standard UKF update step as in \cite{thrun_probabilistic_robotics}.

\subsubsection{Human Wrench Estimation}\label{sec:Wrench_Estimator}
Rearranging eq.~(\ref{eq:total_wrench_on_payload}), we obtain
\begin{equation}
   \begin{bmatrix}\loadmass\loadacc\\\inertiaload\loadangacc\end{bmatrix}=\begin{bmatrix}\extplF \\ \extplM\end{bmatrix}+{\matP}\begin{bmatrix}\tension{1}\\\vdots\\\tension{n}\end{bmatrix}-\begin{bmatrix}
    m\vecg\\
    \mathbf{0}_{3\times1}
    \end{bmatrix}.
    \label{eq:payload_extest}
\end{equation}
Considering quasi-static operating conditions, we can assume the payload linear and angular acceleration terms can be neglected. Therefore, leveraging this assumption and rearranging based on eq.~(\ref{eq:payload_extest}) we obtain
\begin{equation}
\begin{split}
   \begin{bmatrix}\extplF \\ \extplM\end{bmatrix}&=-{\matP}\begin{bmatrix}\tension{1}\\\vdots\\\tension{n}\end{bmatrix}+\begin{bmatrix}
    m\vecg\\
    \mathbf{0}_{3\times1}
    \end{bmatrix}.\label{eq:mapping_tension_external_wrench}
\end{split}
\end{equation}
Extracting tension values from each robot's state allows us to compute the external wrenches on the payload.

\subsection{Payload Admittance Controller} \label{sec:admittance_controller}
The admittance controller is a high-level controller that updates $\loadposdes$ and $\loadrotdes$ in eq.~(\ref{eq:control_errors}) for the payload controller (Section~\ref{sec:payload_controller}) based on the external force applied on the payload. When interacting with the payload, it allows the human operator to experience a virtual mass-spring-damper system rather than the actual mass. By setting the admittance controller's tunable parameters to the desired values, the payload can be either sensitive or insensitive to external forces regardless of the payload's actual property. 

The admittance controller conceptualizes the payload as a virtual mass-spring-damper system, responding to external forces and moments from the human operator. It takes the external wrench, represented as $\begin{bmatrix}\extplF^{\top}&\extplM^{\top}\end{bmatrix}^\top\in\realnum{6\times1}$, as the input. The output from this controller includes the desired payload twist, denoted as $\dot{\mathcal{X}}_{des}\in\realnum{6\times1}$, along with the desired linear and angular positions of the payload, $\mathcal{X}_{des}\in\realnum{6\times1}$, where the angular position is expressed in Euler angles. Additionally, the admittance controller calculates the desired linear and angular accelerations, $\Ddot{\mathcal{X}}_{des}\in\realnum{6\times1}$. However, it is noteworthy that these acceleration outputs are not utilized by the lower-level payload controller.

The admittance controller assumes the following dynamics for the payload
\begin{equation}
\begin{split}
&\mathbf{M}\ddot{\mathbf{e}}_{adm}+\mathbf{D}\dot{\mathbf{e}}_{adm}+\mathbf{K}\mathbf{e}_{adm}=\begin{bmatrix}\extplF\\\extplM\end{bmatrix},\\
&\ddot{\mathbf{e}}_{adm} = \Ddot{\mathcal{X}}_{des}-\Ddot{\mathcal{X}}_{traj},\,\,\,\dot{\mathbf{e}}_{adm} = \dot{\mathcal{X}}_{des}-\dot{\mathcal{X}}_{traj},\\
&\mathbf{e}_{adm} = \mathcal{X}_{des}-\mathcal{X}_{traj},
    \label{eq:admittance_ctrl}
\end{split}
\end{equation}
where $\mathbf{M}$, $\mathbf{D}$, and $\mathbf{K}\in\realnum{6\times6}$ are tunable diagonal positive semidefinite matrices denoting the desired mass, damping, and spring property of the payload.
%, defined as follows:
%\begin{equation}
%\begin{split}
%\mathbf{M} &= diag\prths{M_x,M_y,M_z,M_{\phi},M_{\theta},M_{\psi}},\\
%\mathbf{D} &= diag\prths{D_x,D_y,D_z,D_{\phi},D_{\theta},D_{\psi}},\\
%\mathbf{K} &= diag\prths{K_x,K_y,K_z,K_{\phi},K_{\theta},K_{\psi}},
%\end{split}
%\end{equation}
%where $*_{x,y,z}$ represents parameters corresponding to the $x,y,z$ axes of the translational motion, and $*_{\phi, \theta, \psi}$ denotes parameters associated with the roll, pitch, and yaw directions of the rotational motion, respectively. 
Based on the initial starting condition of the payload, $\Ddot{\mathcal{X}}_{traj}$, $\dot{\mathcal{X}}_{traj}$, and $\mathcal{X}_{traj}$ can be set accordingly. We choose to set them to be the planned trajectory.
Closed-form solutions exist for eq.~(\ref{eq:admittance_ctrl}) with the assumption that the input wrench is a predetermined function (e.g., a linear function). However, such an assumption is not ideal for our use case. Therefore, we choose to solve $\mathcal{X}_{des}$, $\dot{\mathcal{X}}_{des}$, $\Ddot{\mathcal{X}}_{des}$ with Runge-Kutta $4^{th}$ order approximation.

\begin{figure}[t]
    \centering
    \includegraphics[width=\columnwidth]{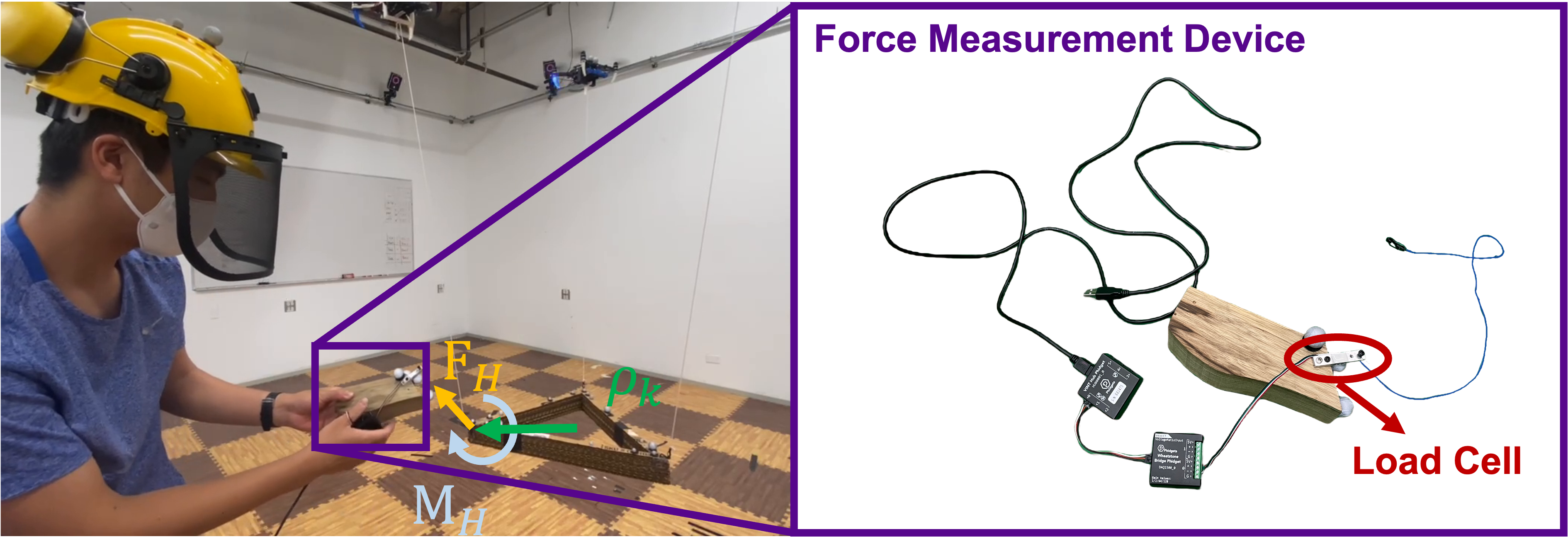}
    \caption{Wrench estimation evaluation. The human operator uses a force measurement device to measure the applied wrench on the payload, which is used to validate our wrench estimation algorithm results. On the left, we show the human operator applies force via the force measurement device, and, on the right, we show the force measurement device in detail.}
    \label{fig:wrench_measurement_device}
\end{figure}
\section{Experimental Results}~\label{sec:experimental_results}
The experiments are conducted in an indoor testbed with a flying space of $10\times6\times4~\si{m^3}$ of the ARPL lab at New York University. We use three quadrotors to carry a triangular payload via suspended cables. The quadrotor platform used in the experiments is equipped with a $\text{Qualcomm}^{\circledR}\text{Snapdragon}^{\text{TM}}$ 801 board for on-board computing~\cite{LoiannoRAL2017}. A laptop equipped with an Intel i9-9900K CPU obtains the Vicon \footnote{\url{www.vicon.com}} motion capture system data via ethernet cable.

The framework has been developed in ROS\footnote{\url{www.ros.org}} and the robots' clocks are synchronized by Chrony\footnote{\url{https://chrony.tuxfamily.org/}}. The mass of the payload is $310~\si{g}$, which exceeds the payload capacity of every single vehicle. The pose and twist of the payload and quadrotors, the position and velocity of attachment points, and the human operator's position are estimated using the Vicon data at a frequency of $100~\si{Hz}$. The unit vector of each cable direction $\cablevec{k}$ and the corresponding velocity $\cabledotvec{k}$ are estimated by
\begin{equation}
    \cablevec{k} = \frac{\aptpos{att,k} - \robotpos{k}}{\norm{\aptpos{att,k} - \robotpos{k}}},~~~\cabledotvec{k} = \frac{\aptvel{att,k} - \robotvel{k}}{l_k},
\end{equation}
where $\aptpos{att,k}, \aptvel{att,k}$ are position and velocity of the $k^{th}$ attach point in $\worldf$ and $\robotpos{k},\robotvel{k}$ are position and velocity of the $k^{th}$ robot in $\worldf$, all of which are estimated by the motion capture system.

\subsection{Cable Force and External Wrench Estimation}
In this section, we validate our cable force and external wrench estimation algorithm by comparing the estimation results obtained using the approach presented in Section~\ref{sec:Wrench_Estimator} with the ground truth from the wrench measurement device as shown in Fig.~\ref{fig:wrench_measurement_device}. 

We can identify the ground truth force by measuring the force direction and force magnitude separately via the wrench measurement device. As the ground truth force direction is along the cable between the measurement device and the other end where the device is attached to the system, it is measured by computing the difference between the load cell's position and the attach point position using the Vicon motion capture system. The ground truth force magnitude is measured via a Phidget micro load cell\footnote{\url{www.phidgets.com}} as shown in Fig.~\ref{fig:wrench_measurement_device}. The measured cable direction and tension magnitude are post-processed to obtain the ground truth force.

\begin{figure}[t]
    \centering
    \includegraphics[width=\columnwidth]{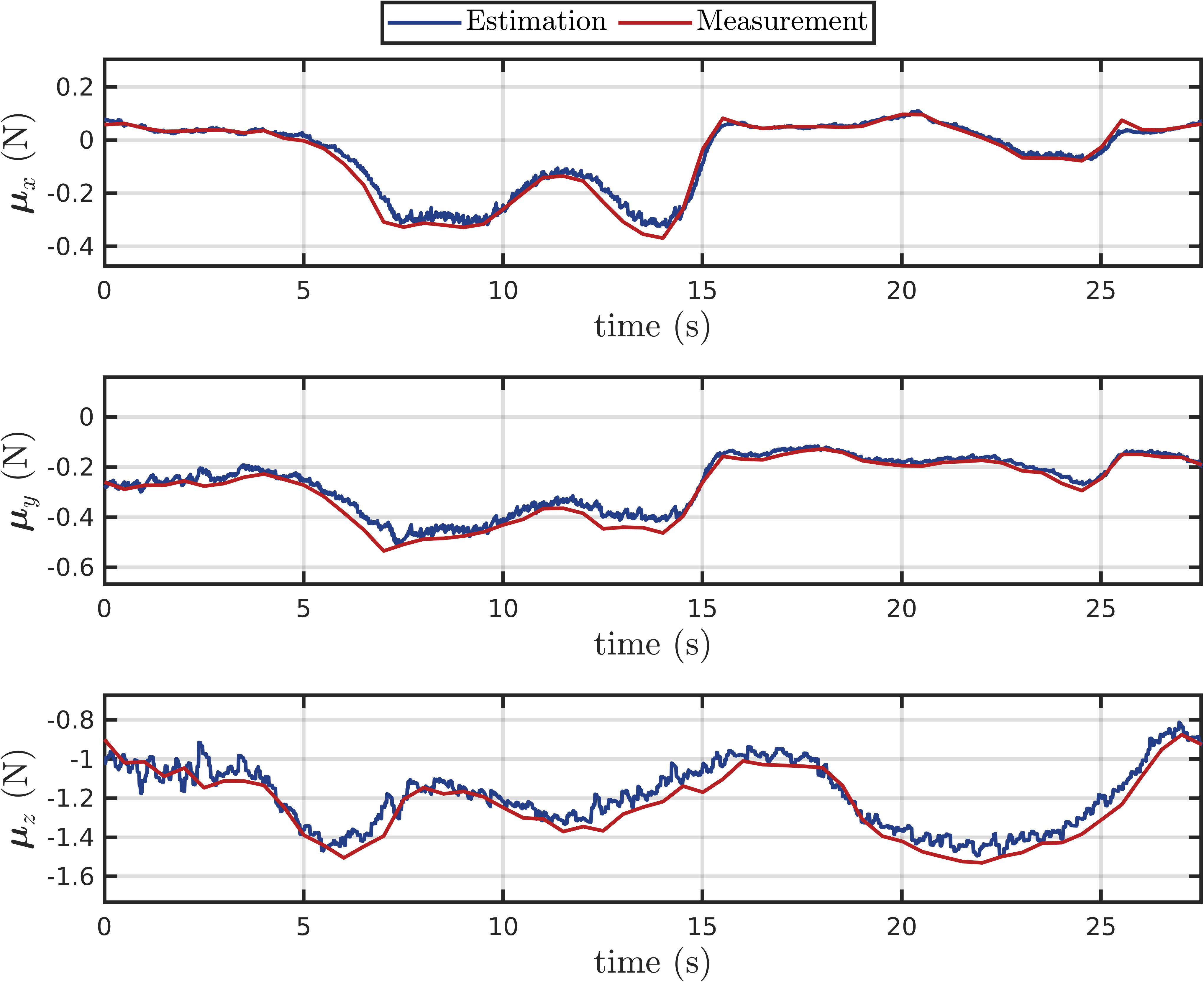}
    \caption{Cable force estimation experiment results. Comparison between the cable force estimation algorithm results and the force measurements from the force measurement device in all 3 DoF.}
    \label{fig:cable_force_measurement}
\end{figure}
\begin{figure*}[t]
    \centering
    \includegraphics[width=\textwidth]{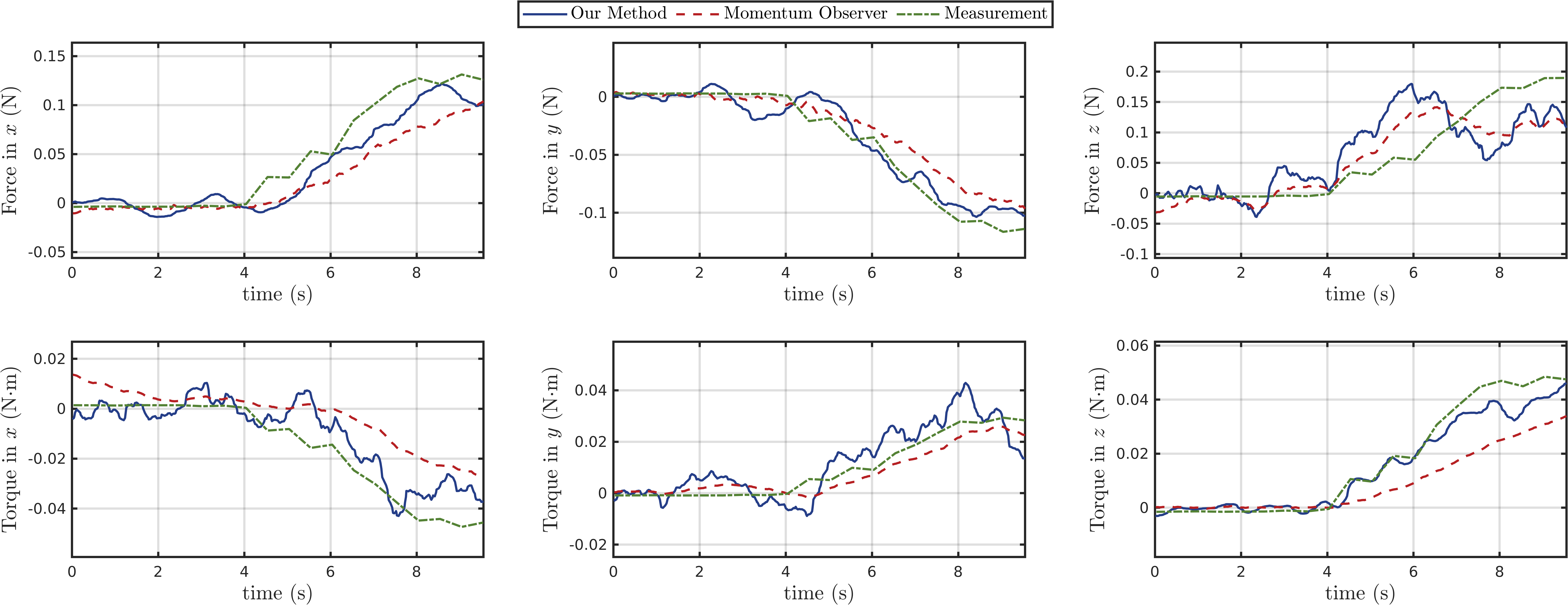}
    \caption{Results of the wrench estimation experiment. This figure compares the wrench estimation results from our proposed wrench estimation algorithm (blue) with those obtained using the momentum observer method (red)~\cite{sanalitro2022ralindirectforce}, as well as with the actual measurements recorded by the wrench measurement device (green) across all 6 DoF.}
    \label{fig:wrench_measurement}
\end{figure*}
\begin{figure*}[t]
    \centering
    \includegraphics[width=\textwidth]{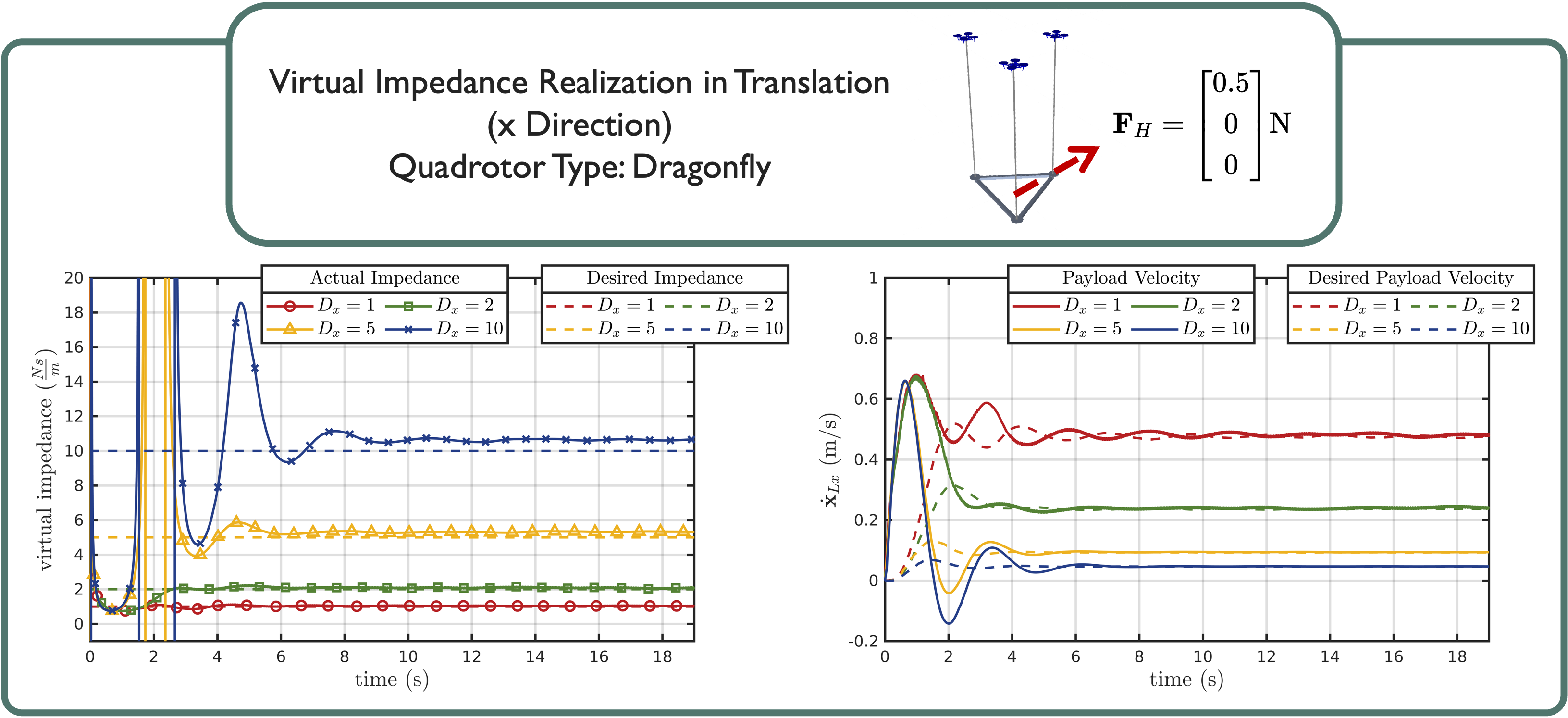}
    \caption{Virtual impedance realization in the $x$ direction of the translational motion. A step force input of $\mathbf{F}_H=\begin{bmatrix}
        0.5&0&0
    \end{bmatrix}^{\top}\si{N}$ is given into the system at the start of the plots. We choose the parameters of the admittance controller in the x direction as $M_x=0.25, K_x = 0.0$, and $D_x=1,~2,~5,~10$ for comparison.}
    \label{fig:dragonfly_virtual_impedance_realization_in_x}
\end{figure*}
\begin{table}[h]
\caption {RMSE of wrench estimation and measurement.\label{tab:wrench_rmse}} 
\centering
%\newcolumntype{s}{}
\begin{tabularx}{\columnwidth}{cccccccc}
\hline\hline
& \multicolumn{3}{c}{Force ($\si{N}$)}& &\multicolumn{3}{c}{Moment ($\si{N\cdot m}$)}\\
  \cline{2-4}\cline{6-8}
&$x$    & $y$ &  $z$ &         &  roll  & pitch &  yaw\\\hline
Ours &0.0185 & 0.0117 & 0.0564& &0.0088 & 0.0066 & 0.0045\\
\cite{sanalitro2022ralindirectforce} &0.0282 & 0.0164 & 0.0419& &0.0148 & 0.0040 & 0.0120\\
\hline\hline
\end{tabularx}
\end{table}

In the cable force estimation experiment, we hover a quadrotor in midair and run the proposed UKF onboard. The measurement device is connected to the center of mass of the quadrotor and a human operator pulls the measurement device into various directions to evaluate the algorithms. The results are shown in Fig.~\ref{fig:cable_force_measurement}. In the plots, we compare the measured forces to the estimated forces in all $3$ DoF and the estimated cable forces track measured ground truth accurately.

During the wrench estimation experiment, we hover the system with the regular payload controller without activating the admittance controller. Subsequently, the human operator pulls the payload with the force measurement device, and we record both the ground truth wrench and the estimation results. In addition to the ground truth force, the ground truth external torque is obtained by crossing the attached point position vector in $\loadf$ and the measured ground truth force vector from the measurement device. The payload is pulled so that the external wrench is non-zero in all six DoF, as shown in Fig.~\ref{fig:wrench_measurement_device}. 

\begin{figure*}[t]
    \centering
    \includegraphics[width=\textwidth]{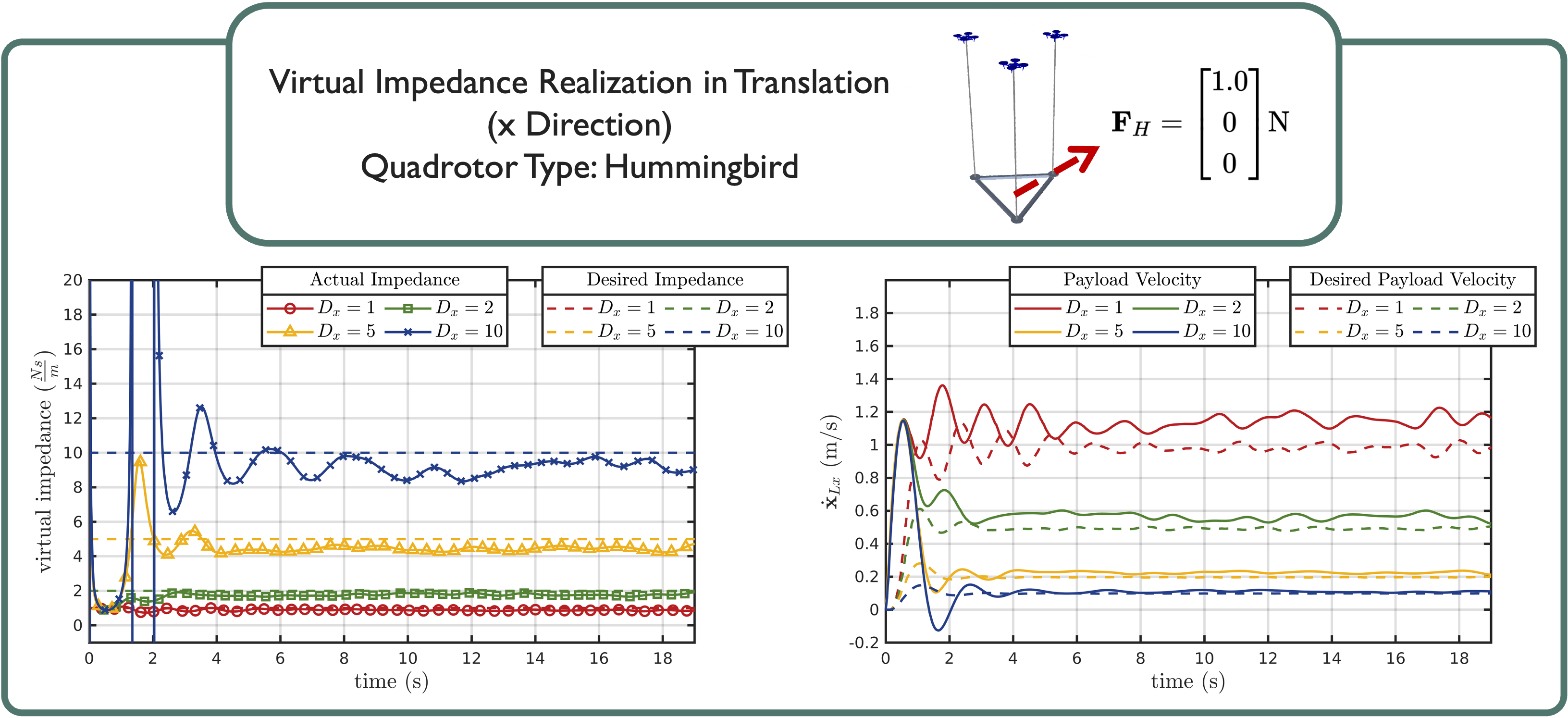}
    \caption{Virtual impedance realization in the $x$ direction of the translational motion. A step force input of $\mathbf{F}_H=\begin{bmatrix}
        1.0&0&0
    \end{bmatrix}^{\top}\si{N}$ is given into the system at the start of the plots. We choose the parameters of the admittance controller in the x direction as $M_x=0.25, K_x = 0.0$, and $D_x=1,~2,~5,~10$ for comparison.}
    \label{fig:hummingbird_virtual_impedance_realization_in_x}
\end{figure*}
\begin{figure*}[!t]
    \centering
    \includegraphics[width=\textwidth]{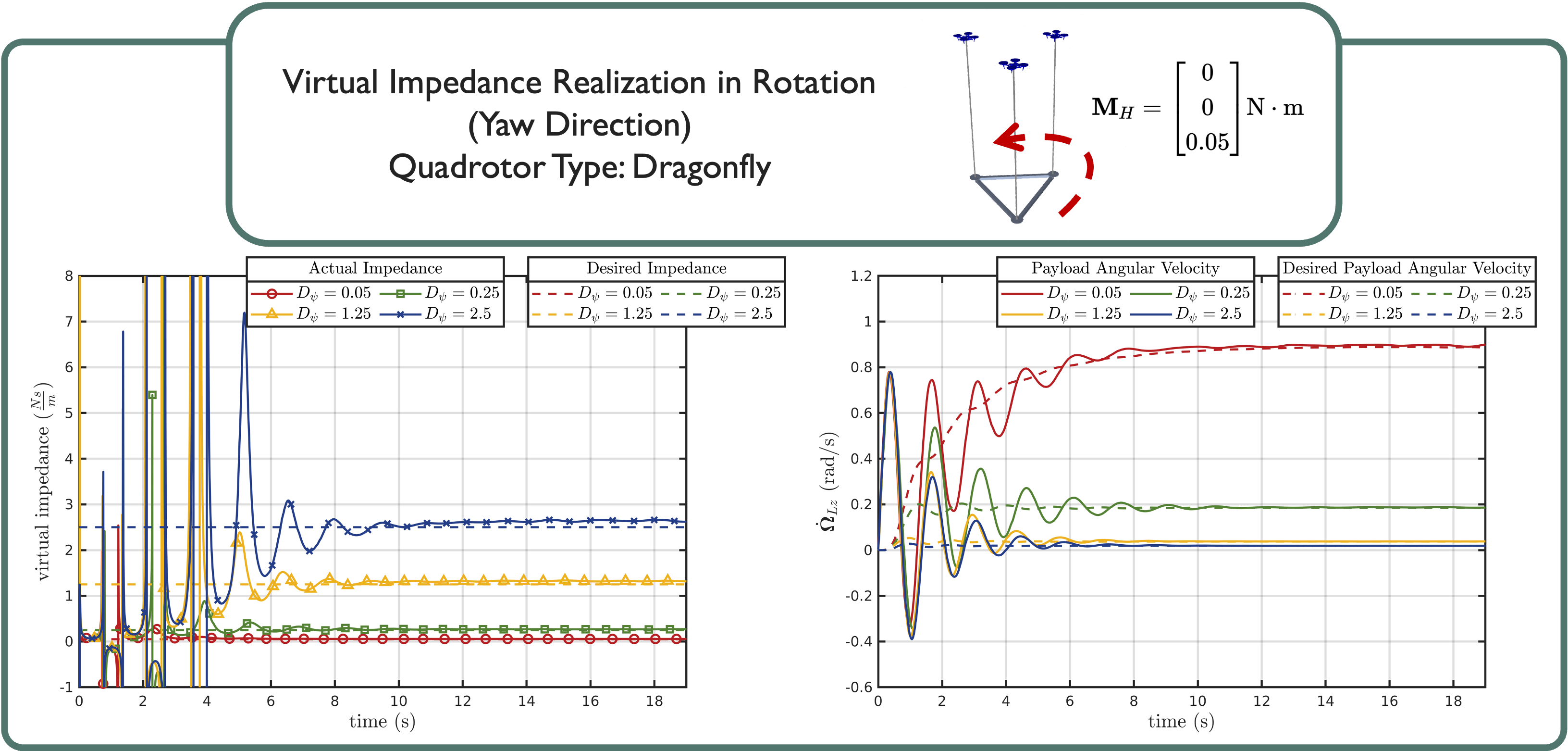}
    \caption{Virtual impedance realization in the yaw direction of the rotational motion. A step moment input of $\mathbf{M}_H=\begin{bmatrix}
        0&0&0.05
    \end{bmatrix}^{\top}\si{Nm}$ is given into the system at the start of the plots. We choose the parameters of the admittance controller in the yaw direction as $M_{\psi}=0.1,~K_{\psi} = 0.0$, and $D_{\psi}=0.05,~0.25,~1.25,~2.5$ for comparison. }
    \label{fig:virtual_impedance_realization_in_yaw}
\end{figure*}

The results are shown in Fig.~\ref{fig:wrench_measurement}. In the plots, we compare the measured wrenches to the estimated wrenches using our proposed method in Section~\ref{sec:Wrench_Estimator} and the momentum observer method presented in~\cite{sanalitro2022ralindirectforce} in all 6 DoF. As we can observe in Fig.~\ref{fig:wrench_measurement}, the estimated wrenches from our method track the measured ground truth quite accurately. However, on the other hand, the momentum observer method tends to smooth the estimates excessively, leading to underestimating the external wrench. The root mean square errors in all six directions are also reported in Table \ref{tab:wrench_rmse}, confirming a good accuracy. 

\begin{figure*}[!t]
\centering
  \includegraphics[width=\textwidth]{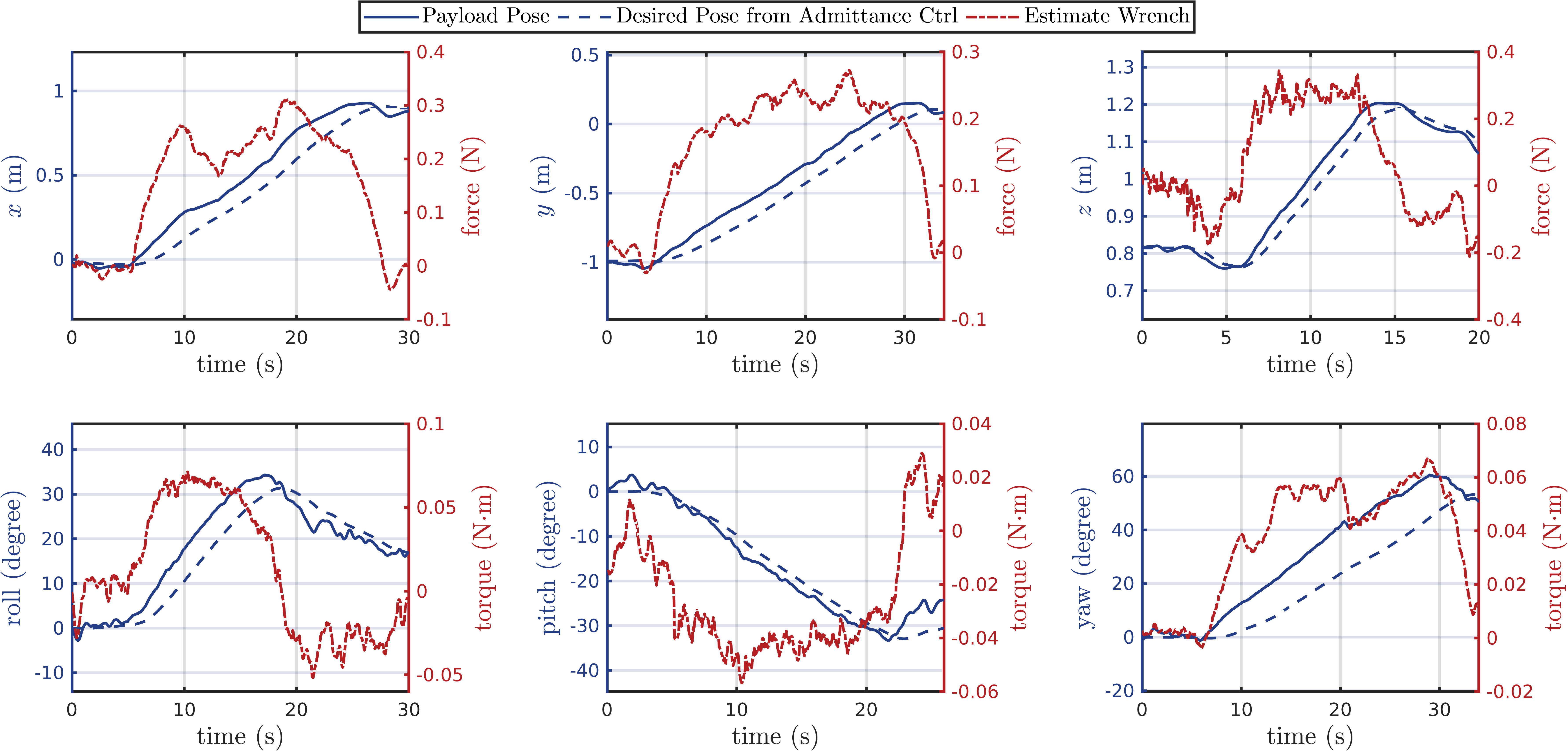}  \caption{We test the admittance controller together with wrench estimation in all 6 DoF. The $\mathcal{X}_{traj}$ are where the plots start and the derivative $\dot{\mathcal{X}}_{traj}$ $\ddot{\mathcal{X}}_{traj}$ are zero. The actual payload pose, desired payload pose from the admittance controller, and estimated wrench are plotted. \textit{On the top:} translation tests in $x$, $y$, $z$ in $\worldf$. \textit{On the bottom:} rotation tests in roll, pitch, yaw in $\loadf$. \label{fig:admittance_control_test_results}}
\end{figure*}

\subsection{Admittance Control with Wrench Estimation}
After validating the wrench estimation, we jointly test it with the admittance controller.
\subsubsection{Virtual Impedance Realization}
In this section, we present results in simulation to quantitatively analyze the performance of our proposed methods, particularly regarding the rendering of the desired virtual impedance of our proposed system. We deploy robot teams that consist of $3$ ``Dragonfly" quadrotors or $3$ ``Hummingbird" quadrotors with $1~\si{m}$ cable in our open-source simulator~\cite{guanrui2022rotortm} to validate the realization of desired impedance values. In the following experiments, we introduce a step wrench input and observe the system's response. 

In the first experiment, we apply a step force input on the payload in the positive $x$ direction and set different impedance values ($1$,~$2$,~$5$,~$10$) in the admittance controller.  The second experiment involves the application of a step moment input on the payload in the positive yaw direction, with different impedance values ($0.05$, $0.25$, $1.25$, $2.5$) set in the admittance controller. The results are shown in Figs.~\ref{fig:dragonfly_virtual_impedance_realization_in_x}- \ref{fig:virtual_impedance_realization_in_yaw}. We evaluate the actual impedance in the system using the ground truth force and moment, divided by the actual velocity of the payload along the corresponding direction. Hence the actual impedance we obtain in Figs.~\ref{fig:dragonfly_virtual_impedance_realization_in_x}- Fig.~\ref{fig:virtual_impedance_realization_in_yaw} are $\frac{\vecF_{Hx}}{\dot{\vecx}_{Lx}}$ and $\frac{\matM_{Hz}}{\dot{\bm{\Omega}}_{Lz}}$, respectively. In the plots, we compare the actual impedance with the desired impedance set in the admittance controller.  

As illustrated by the plots, upon application of step force and moment inputs to the system, the payload promptly accelerates in the $x$ and yaw directions, respectively. As the wrench estimation updates, the admittance controller starts to adjust the desired payload state to adapt the human input, shown by the dashed lines in the bottom plots in both Figs.~\ref{fig:dragonfly_virtual_impedance_realization_in_x} - \ref{fig:virtual_impedance_realization_in_yaw}. Through the comparison, we can see that with larger values of $D_x, D_{\psi}$, the desired velocity derived from the admittance controller evolves slower and smaller in magnitude. As the experiment proceeds, the desired velocity from the admittance controller ultimately converges to final values, respectively equivalent to $\frac{\vecF_{Hx}}{D_{x}}$ and $\frac{\matM_{Hz}}{D_{\psi}}$.

Moreover, a larger desired impedance value results in noticeable spikes in the actual impedance before the convergence. This can be attributed to the fact that a higher impedance value leads to a smaller corresponding desired velocity, which in turn causes larger velocity errors at the initial stage. This subsequently results in an overshooting response of the actual payload velocity, causing it to cross the zero line and trigger spikes in the actual impedance realization.

Lastly, as the experiments progress and the transient effects resulting from the step input reduce, the payload's linear and angular velocities converge toward the desired payload velocity determined by the admittance controller. Consequently, the actual virtual impedance also aligns with the value set by the admittance controller. We observe similar results for the other Cartesian and angular axes that are not reported for simplicity.

\begin{figure}[t]
\centering
  \includegraphics[width=\columnwidth]{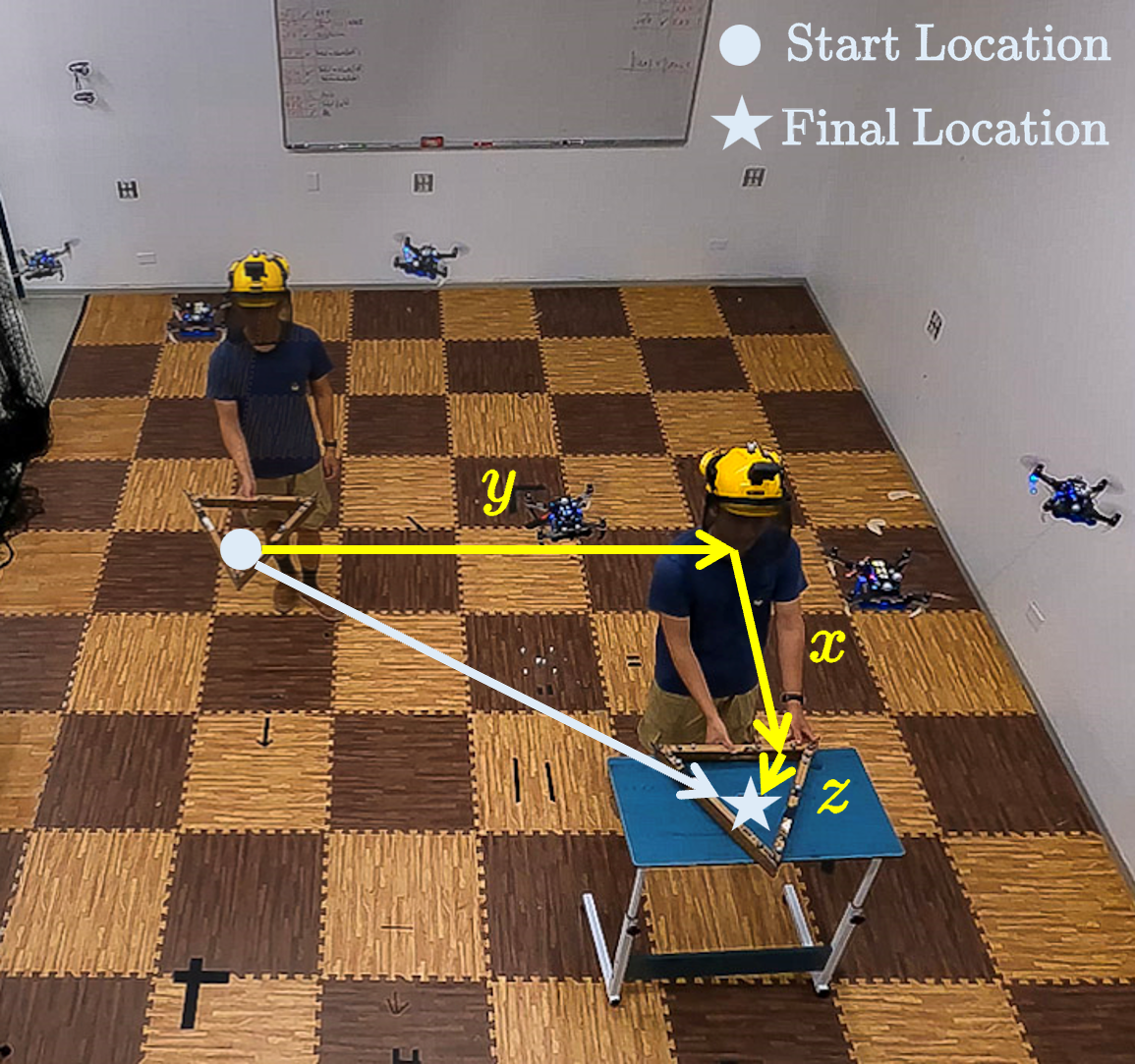}  
  \caption{Human-robot collaborative transportation task. In this task, the human operator collaborates with a team of quadrotors to transport a payload from the start position ($\bullet$) to the final position ($\bigstar$). The human operator and the quadrotor team translate the payload in the Cartesian space, moving its position along the \(x\), \(y\), and \(z\) axes. \label{fig:start2goal}}
\end{figure}
\begin{figure}[t]
    \centering
    \includegraphics[width=\columnwidth]{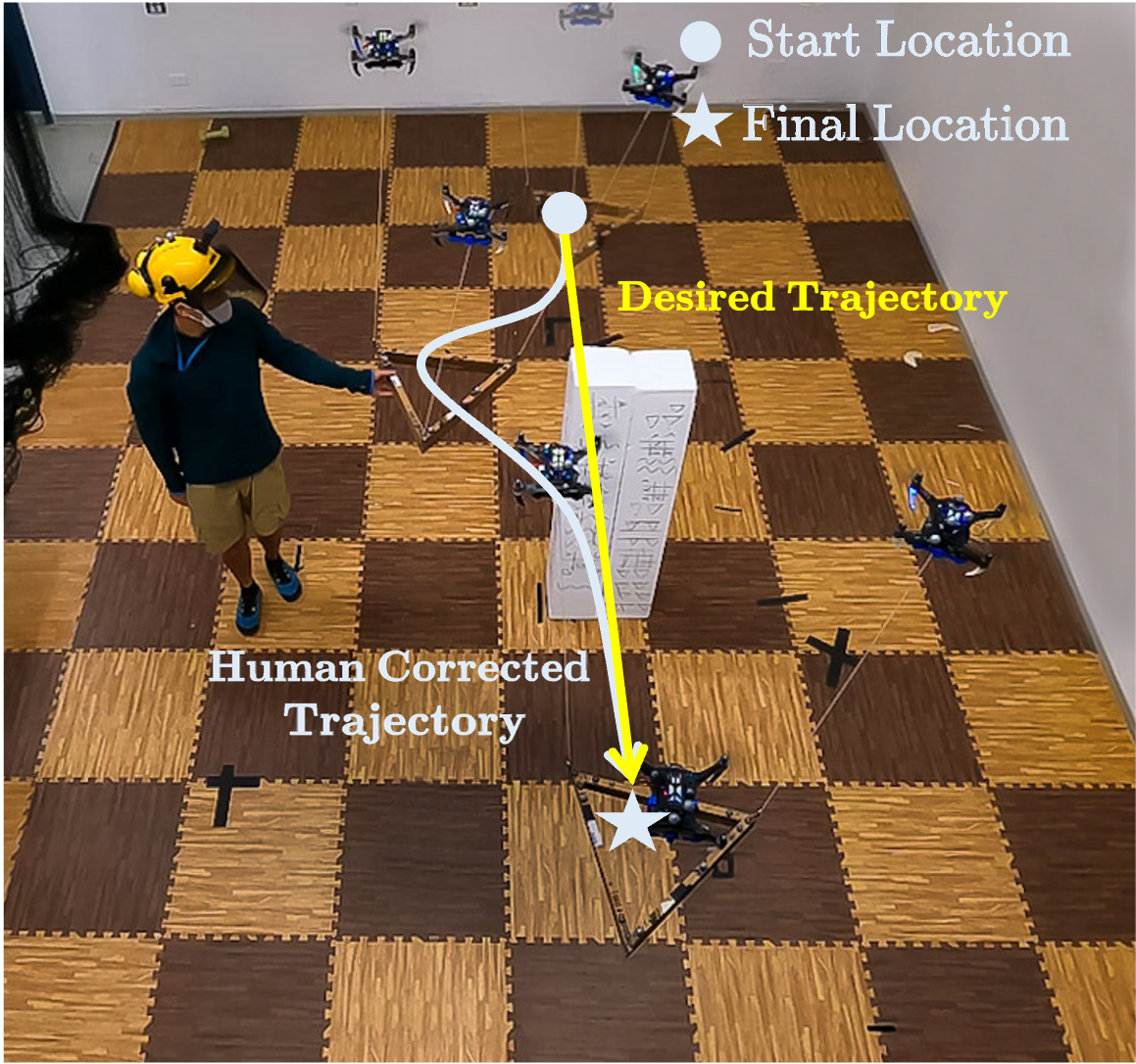}
    \caption{Human-assisted obstacle avoidance task. In this task, the human operator moves the payload from the desired trajectory (yellow path) to guide the payload away from an unknown obstacle and then releases the payload, allowing it to rejoin the desired trajectory and reach the final position ($\bigstar$).}
         \label{fig:avoid_obstacle_task}
\end{figure}
\begin{figure*}[t]
\centering
  \includegraphics[width=\textwidth]{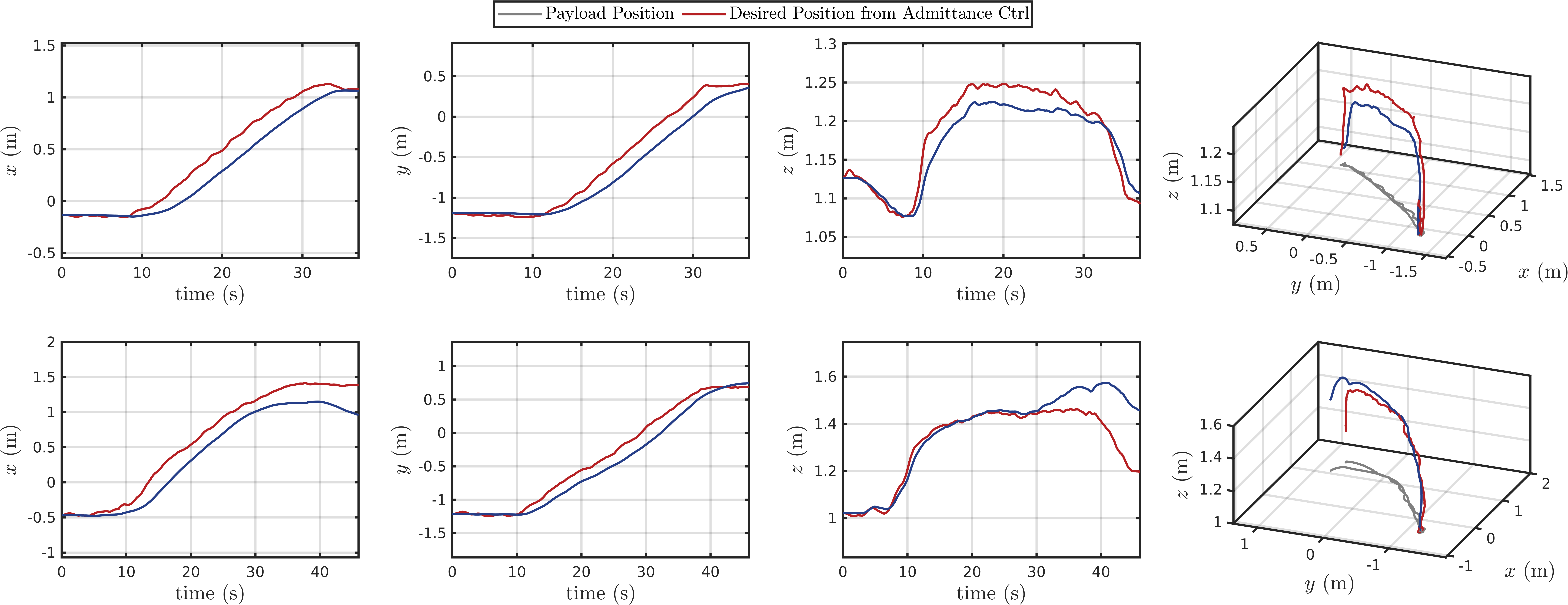}  \caption{Human-robot collaborative transportation experiment results. Payload Position vs. Desired Position from Admittance Controller. Transnational results when optimization-based human-aware method is used (top row). Transnational results when gradient-based human-aware method is used (bottom row).\label{fig:Task1result}}
\end{figure*}
\begin{figure*}[t]
\centering
  \includegraphics[width=\textwidth]{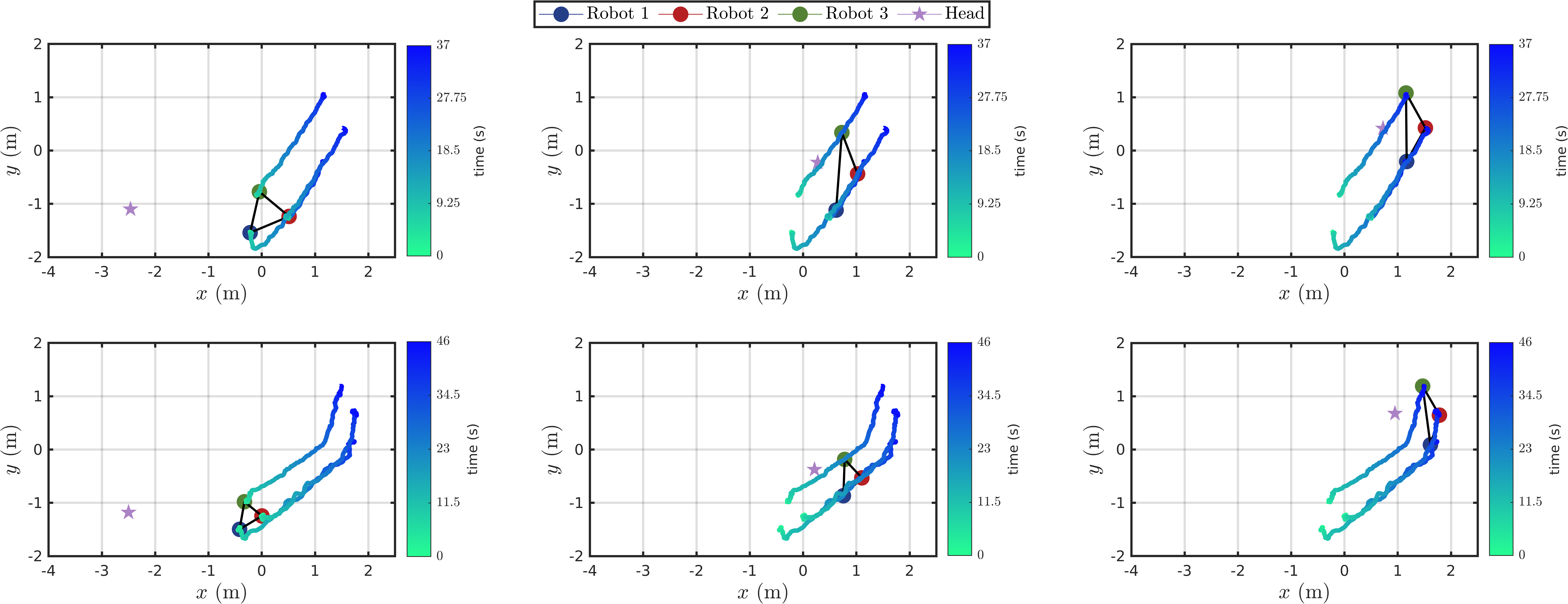}  \caption{Human-robot collaborative transportation experiment results. We show the top down view of the human operator (star) and the team of drones (circles). \textit{On the top:} trajectory when optimization-based human-aware method is used. \textit{On the bottom:} trajectory when gradient-based human-aware method is used.\label{fig:human_guided_transportation_robot_formation}}
\end{figure*}
\begin{figure}[t]
    \centering
    \includegraphics[width=\columnwidth]{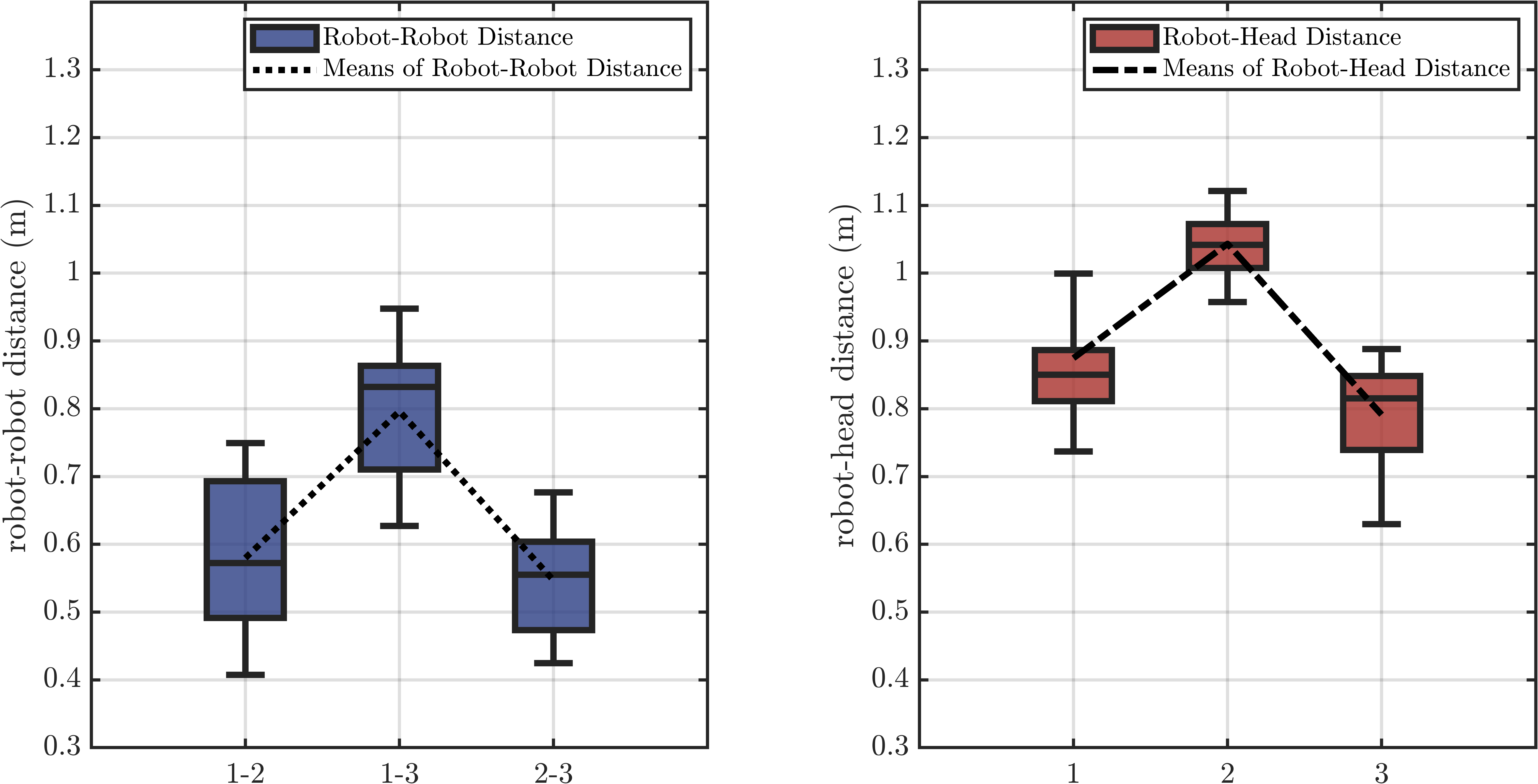}
    \caption{Box plot with mean lines of the robot to robot distances and robot to head distances during the human-aware collaborative transportation experiments using the gradient-based method. The data are collected from multiple repeated experiments.}
         \label{fig:gradient-analysis}
\end{figure}
\begin{figure}[t]
    \centering
    \includegraphics[width=\columnwidth]{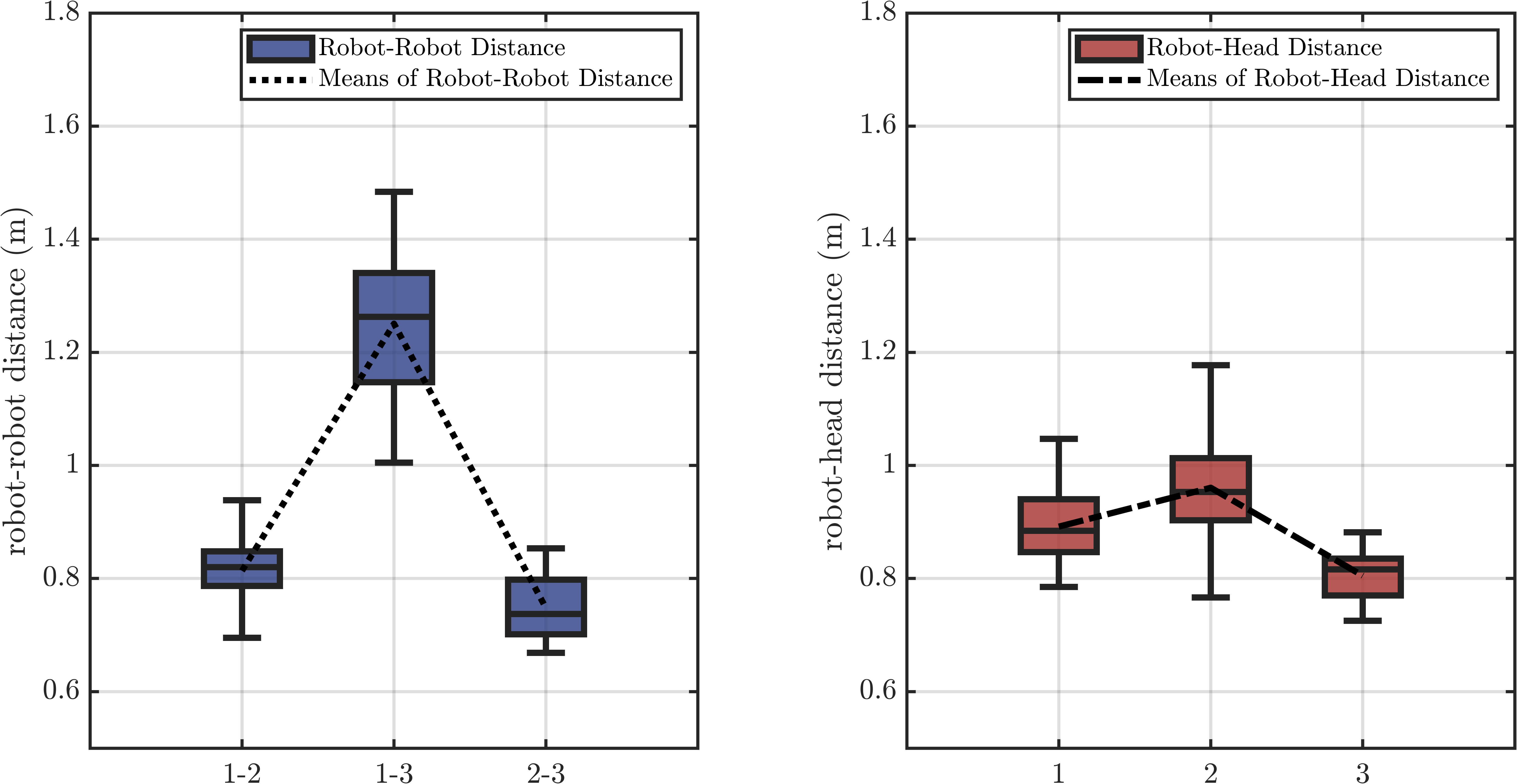}
    \caption{Box plot with mean lines of the robot to robot distances and robot to head distances during the human-aware collaborative transportation experiments using the optimization-based method. The data are collected from multiple repeated experiments.}
    \vspace{-20pt}
         \label{fig:optimization-analysis}
\end{figure}
\begin{figure}[t]
    \centering
    \includegraphics[width=\columnwidth]{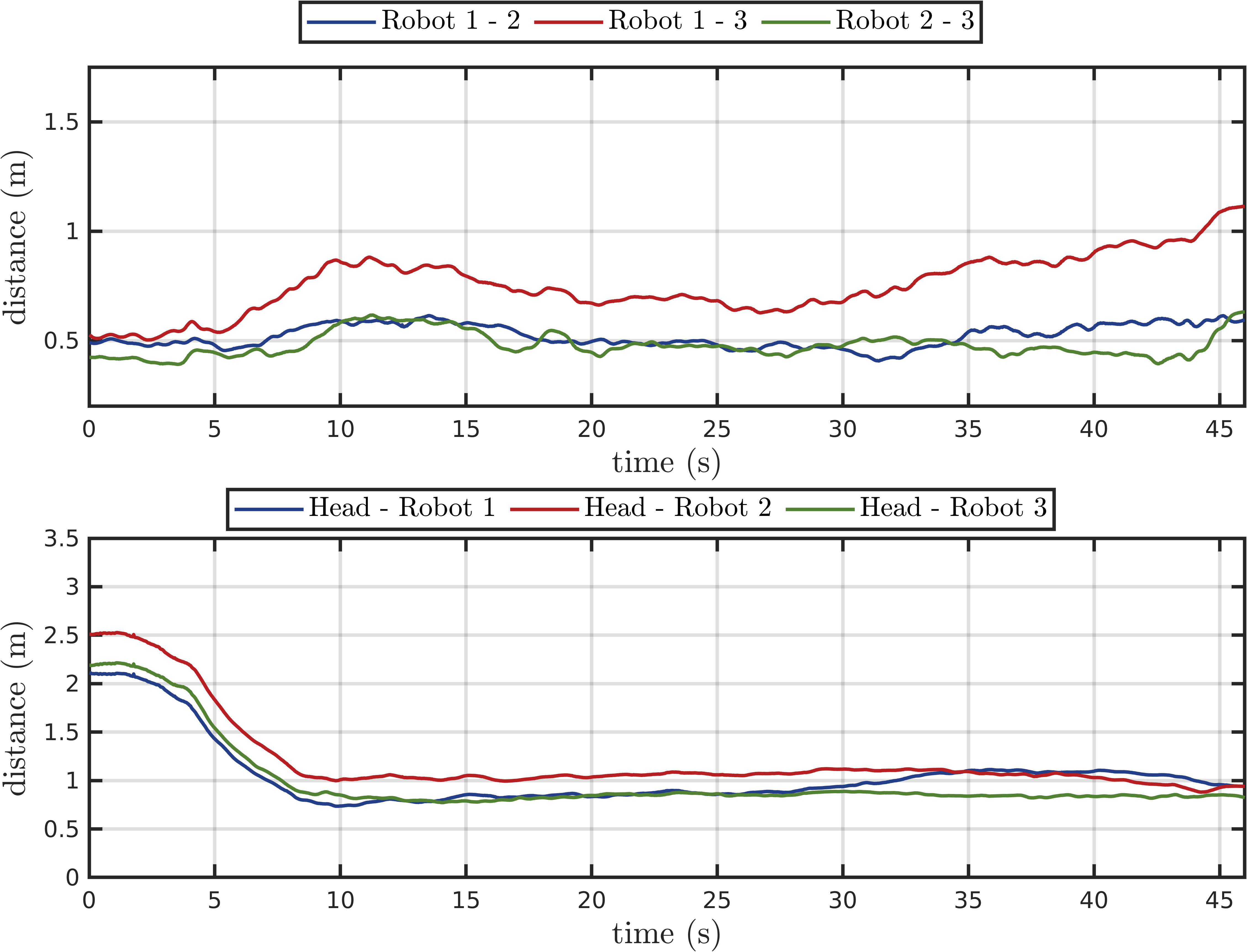}
    \caption{Human-robot collaborative transportation experiment results. Drone to drone distance and human to drone distance when gradient-based method is used; the gradient-based method does not require a predetermined distance.}
    \label{fig:transportation_distance_grad}
\end{figure}
\begin{figure}[t]
    \centering
    \includegraphics[width=\columnwidth]{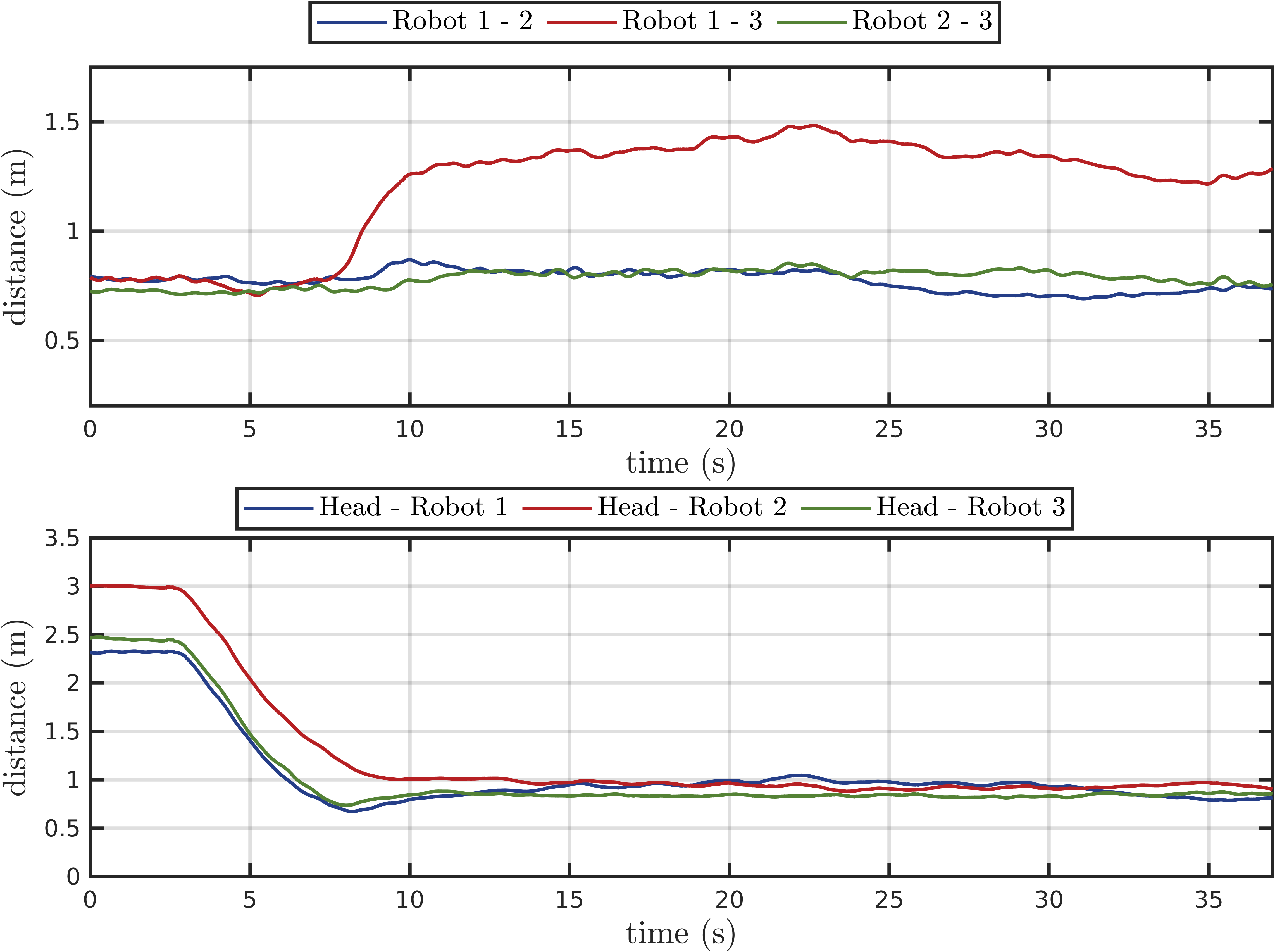}
    \caption{Human-robot collaborative transportation experiment results. Drone-to-drone distance and human-to-drone distance when optimization-based method is used; minimum drone-to-drone distance is set to be 0.75 m, and the minimum human-to-drone distance is set to be 0.75 m.}
    \label{fig:transportation_distance_opt}
\end{figure}

\subsubsection{Real-World Experiments}
 We conduct six tests involving all $6$ DoF of the admittance controller in real-world experiments. The human operator manipulates the payload by translating the payload in $x$, $y$, and $z$ and rotating the payload in roll, pitch, and yaw, respectively, to show that the load can be fully manipulated. At the end of each experiment, the human operator releases the payload. The square gain matrices in the admittance controller have a block-diagonal structure as $
  \mathbf{M} = diag(0.25\mathbf{I}_{3\times3}, 0.1\mathbf{I}_{3\times3}),
  ~\mathbf{D}  = diag(1.25\mathbf{I}_{3\times3}, 5.0\mathbf{I}_{3\times3}),~
  \mathbf{K} =diag(\mathbf{0}_{3\times3}, \mathbf{0}_{3\times3})
$.

The experimental results are presented in Fig.~\ref{fig:admittance_control_test_results}. As shown in the plots, the human operator translates the payload approximately $1$ m in $x$ and $y$ direction, $0.4$ m in the $z$ direction. In the rotation part of the experiment, the human operator rotates the payload approximately $30^\circ$ in the roll and pitch direction and $60^\circ$ in the yaw direction. The tests show that the admittance controller, coupled with the wrench estimator, can successfully update the payload's desired position or orientation according to the human operator's interactive force as input.
As the human operator releases the payload, the wrench estimation outputs $\begin{bmatrix}\mathbf{0}_{6\times1}\end{bmatrix}$ as wrench estimation. Since $\mathbf{K}$ in the admittance controller is $\begin{bmatrix}\mathbf{0}_{6\times6}\end{bmatrix}$ in this set of experiments, the payload remains at the position or orientation released by the human operator without returning to its original reference position or orientation. It further confirms the effectiveness of the wrench estimation and admittance controller pipeline in assisting object transportation and manipulation.

\subsection{Human-Aware Human-Robot Collaborative Transportation}
In this section, we show that our system enables a human operator to physically collaborate with the robot team to accomplish the following two tasks
\begin{enumerate}
    \item The robot team and the human operator collaboratively manipulate the payload to a goal location, as demonstrated in Fig.~\ref{fig:start2goal}.
    \item Human operator corrects the payload trajectory to avoid an obstacle in an existing trajectory, as demonstrated in Fig.~\ref{fig:avoid_obstacle_task}
\end{enumerate}
The gradient-based and optimization-based methods for human-aware force distribution are tested for each of the two tasks. For the optimization-based method, the drone-to-drone distance limit is set to be $\geq 0.75~\si{m}$, and the human-to-drone distance limit is set to be $\geq 0.75~\si{m}$. The gradient-based method does not require a predetermined distance.

In addition, as we show in Fig.~\ref{fig:control_block_diagram} and discuss in Section~\ref{sec:force_allocation}, we feed the human operator’s position $\Phuman$ from the Vicon in $\worldf$ as $\Pobject$ to the eqs.~(\ref{eq:robotkgrad}) and (\ref{eq:nlopt}) for application to physical human-robot interaction. We would also like to note that by using deep-learning-based human pose estimation techniques~\cite{rennips2015fasterrcnh}, the robots can also use onboard camera to estimate $\Phuman$, but this is out of the scope of this paper and we refer to it as future work. 

\subsubsection{Human-robot collaborative transportation}
In this experiment, the robot team and the human operator collaborate together to move a payload from the starting location to the final location via direct force interaction. Payload translates in all three axes, as shown in Fig. \ref{fig:start2goal}. The square gain matrices are selected block-diagonal as $
  \mathbf{M} = diag(0.25\mathbf{I}_{3\times3}, 0.1\mathbf{I}_{3\times3}),
  ~\mathbf{D}  = diag(5.0\mathbf{I}_{3\times3}, 5.0\mathbf{I}_{3\times3}),~
  \mathbf{K} =diag(\mathbf{0}_{3\times3}, \mathbf{0}_{3\times3})
$.

Note that the spring constant coefficient for the admittance controller is set to zero so that the payload stays at the position/orientation once the human operator releases the payload.

The experimental results are shown in Fig.~\ref{fig:Task1result}, where we compare the actual payload position with the desired payload position from the admittance controller. 

The results demonstrate the proposed methods can confidently update the desired payload position to satisfy the human operator's intention of moving the payload under both gradient-based and optimization-based methods. Furthermore, the movement introduced by the human-aware force distribution does not affect the performance of the payload wrench estimation or admittance controller. 

In Fig.~\ref{fig:human_guided_transportation_robot_formation}, we show the effects of the two methods for human-aware force distribution with a top view of the entire collaboration task. From the plots, we can observe that, as the human operator, denoted by the purple star, approaches the $3$ robots with a suspended payload, the human-aware force distribution starts to be effective. The controller expands the $2$ robots (blue and green circles) that are close to the human operator to keep the distance.

\begin{table}[t]
\caption {Distance Summary: Robot-Robot and Robot-Head Pairs.\label{tab:distance_summary}} 
\centering
\begin{tabularx}{0.8\columnwidth}{c|ccc}
    \hline\hline
    Method & Pair Type & Mean & Std \\
    \hline
    \multirow{6}{*}{Gradient-based}
      & 1-2 & 0.57978 & 0.096378 \\
      & 1-3 & 0.79629 & 0.084538 \\
      & 2-3 & 0.54679 & 0.076384 \\
      & Head-1   & 0.87464 & 0.097718 \\
      & Head-2   & 1.0433  & 0.041136 \\
      & Head-3   & 0.79129 & 0.071813 \\
    \hline
    \multirow{6}{*}{Optimization-based}
      & 1-2 & 0.81366 & 0.053177 \\
      & 1-3 & 1.2506  & 0.10804  \\
      & 2-3 & 0.74762 & 0.050086 \\
      & Head-1   & 0.89176 & 0.05772  \\
      & Head-2   & 0.96053 & 0.080423 \\
      & Head-3   & 0.80493 & 0.038652 \\
    \hline\hline
\end{tabularx}
\end{table}
To quantitatively support our analysis, we conducted multiple iterations of the same experiments and recorded the distances between the robots and the human operator. The distribution of these distance measurements is illustrated in Figs.~\ref{fig:gradient-analysis} and \ref{fig:optimization-analysis}, with corresponding mean and standard deviation values summarized in Table~\ref{tab:distance_summary}.

As shown in Figs.~\ref{fig:gradient-analysis} and \ref{fig:optimization-analysis}, both the optimization-based and gradient-based methods effectively maintain a consistent distance between the robots and the human operator throughout multiple repeated experiments. This observation is further validated by the statistical data presented in Table~\ref{tab:distance_summary}. Notably, the optimization-based method distinguishes itself from the gradient-based method as it enforces inter-robot distance constraints. This is evident in the left plot of Fig.~\ref{fig:optimization-analysis}, where all three drones maintain a minimum separation of $0.75\si{m}$, as specified by the constraint.

To provide additional insights, Figs.~\ref{fig:transportation_distance_grad} and \ref{fig:transportation_distance_opt} depict the distances between each robot and the human operator, as well as the inter-robot distances, throughout the duration of a single sample experiment. Initially, the human operator starts approximately $2$ to $3\si{m}$ away from the robot team. As the operator approaches, the distances between the human and robots 1 and 3 (represented by blue and green) decrease. At this point, the human-aware force distribution becomes active, maintaining stable human-robot distances.

\begin{table*}[t]
    \centering
    \caption{Computational Complexity Summary}
    \begin{tabular}[\textwidth]{ccc}
    \hline\hline
        &   Scalability & Time Complexity\\
        \hline
\cellcolor{mygreen}Physical Human-Robot Interaction&\cellcolor{mygreen}&\cellcolor{mygreen}\\\hline
      Robot State Estimation&  High & $\mathcal{O}(1)$\\
      Human Wrench Estimation& Medium  & $\mathcal{O}(n)$\\
      Payload Admittance Controller&  High & $O(1)$\\
      \hline
\cellcolor{myblue}Planning and Control&\cellcolor{myblue}&\cellcolor{myblue}\\\hline
      Payload Trajectory Planner & High  & $\mathcal{O}(1)$\\
      Payload Trajectory Tracking Controller & High  & $\mathcal{O}(1)$\\
      Nominal Force Distribution& Medium   & $\mathcal{O}(n)$\\
      Human-Aware Force Distribution: Gradient-Based Method& High  & $\mathcal{O}(n)$\\
      Human-Aware Force Distribution: Optimization-Based Method\hspace{10em}&  Low  & $\mathcal{O}(n^3)$\\
      Robot Controller& High  & $\mathcal{O}(1)$\\
      \hline\hline
    \end{tabular}
    \vspace{-20pt}
    \label{tab:complexity}
\end{table*}
\subsubsection{Human-assisted Obstacle Avoidance}
In this experiment, the payload follows a straight trajectory from the starting location to the final location, as the robot team is unaware of the obstacle. The human operator corrects the payload trajectory to avoid the obstacle, as shown in Fig.~\ref{fig:avoid_obstacle_task}. Both gradient-based and optimization-based methods are also applied here. The square gain matrices for the admittance controller have a block-diagonal structure as $
  \mathbf{M} = diag(0.25\mathbf{I}_{3\times3}, 0.1\mathbf{I}_{3\times3}),
  ~\mathbf{D}  = diag(5.0\mathbf{I}_{3\times3}, 5.0\mathbf{I}_{3\times3}),~
  \mathbf{K} =diag(1.2\mathbf{I}_{3\times3}, \mathbf{0}_{3\times3})
$.

Note that the constant spring coefficient for the admittance controller is no longer zero. The payload will now return to the position/orientation commanded by the trajectory when the human operator releases the payload.

\begin{figure}[t]
    \centering
    \includegraphics[width=\columnwidth]{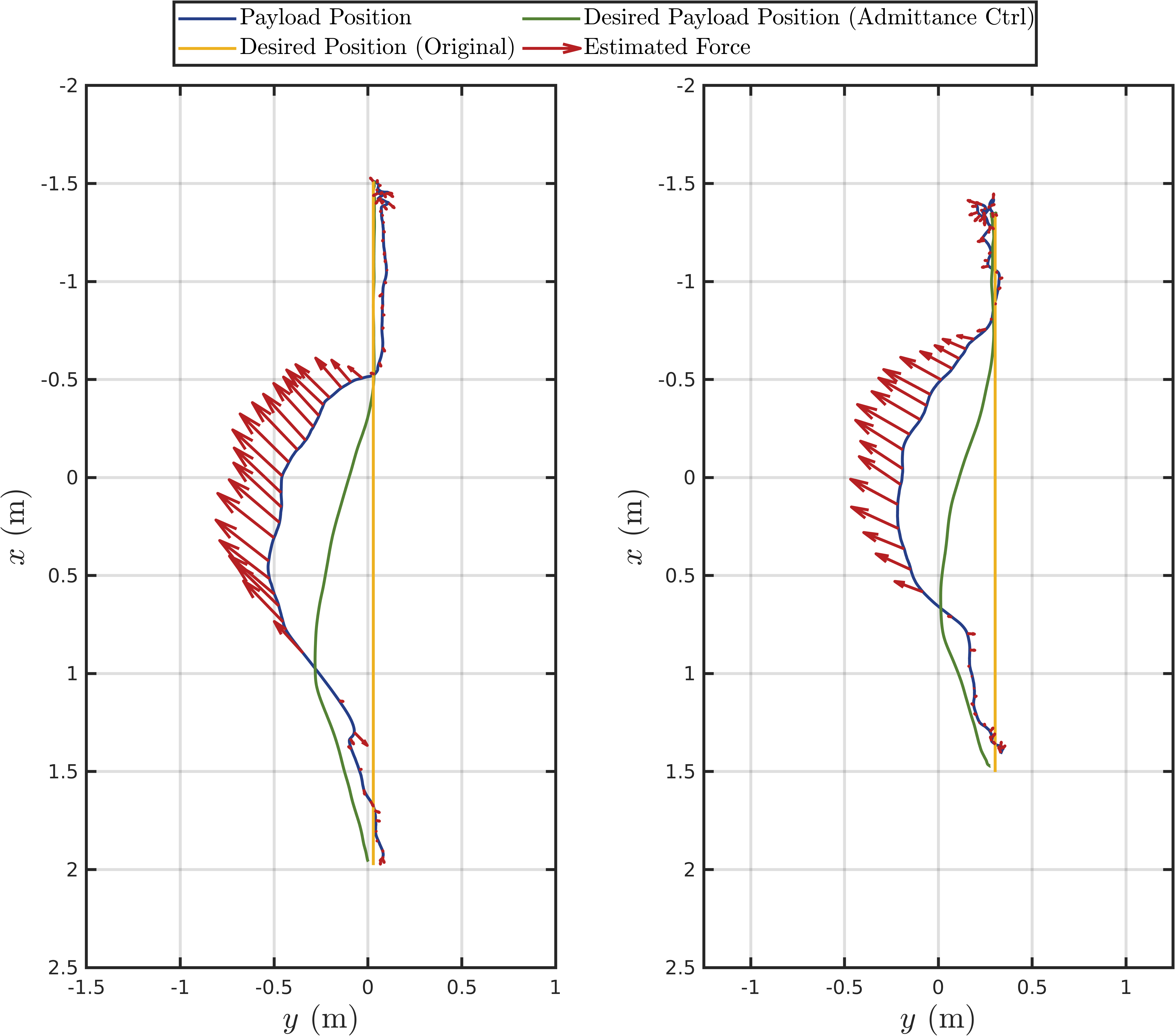}
    \caption{Correcting payload trajectory experiment result. Comparison between actual trajectory, desired trajectory, admittance controller output, and estimated external wrench on the payload. Optimization-based safety controller is used (left). Gradient-based safety controller is used (right).}
         \label{fig:avoid_obstacle_task_plot}
\end{figure}

As we can see from Fig.~\ref{fig:avoid_obstacle_task_plot}, the correction takes effects according to the admittance controlled trajectory. Once the human operator stops the correction, the non-zero $\mathbf{K}$ constant starts to allow the corrected trajectory to converge with the original trajectory. As expected, such behavior is present in both the gradient-based and optimization-based methods.

\section{Computational Complexity Discussion}~\label{sec:discussion}
In this section, we discuss the theoretical computational time complexity of the methods proposed in this paper, focusing on how each algorithm's computational complexity scales with the number of robots. Through this discussion, we aim to offer insights into the proposed methods and guide the corresponding system design choices. We summarize the results regarding the computational complexity of our algorithms in Table~\ref{tab:complexity}.

\subsection{Physical Human-Robot Interaction}
We begin our discussion with the Physical Human-Robot Interaction module, comprising the robot state estimator, the human wrench estimator, and the payload admittance controller.

Firstly, each quadrotor runs its robot state estimator, a UKF with a fixed state vector size of $19$ and a fixed input vector size of $4$. Similarly, the payload admittance controller's computation, as described in eq.~\eqref{eq:admittance_ctrl}, is independent of the number of quadrotors, as it controls only the $6$ degrees of freedom (DoFs) of the payload. Consequently, both the UKF and the payload admittance controller exhibit a computational complexity of $\mathcal{O}(1)$, independent of the number of quadrotors $n$ in the system.

Next, the human wrench estimation requires quadrotors in the team to share individual estimated cable tension forces, which are then aggregated to derive the total external wrench on the payload using eq.\eqref{eq:mapping_tension_external_wrench}. The computation, shown in eq.\eqref{eq:mapping_tension_external_wrench}, scales linearly with the number of quadrotors due to the linear tension mapping matrix $\matP$. Thus, the computation complexity in eq.\eqref{eq:mapping_tension_external_wrench} is bounded by $\mathcal{O}(n)$, with the dimension of matrix $\matP$, as depicted in eq.\eqref{eq:P_mat}, scaling accordingly with $n$. This linear time complexity is manageable with the available computational resources, as $n$ would need to reach the order of thousands to make this linear-complexity matrix multiplication the system's bottleneck.

\subsection{Planning and Control}
In the domain of planning and control, the system includes payload trajectory planner, payload trajectory tracking controller, dynamic force distribution, and robot controller. The payload trajectory planner and tracking controller function similarly to the payload admittance controller, meaning their computation also remains independent of the number of quadrotors $n$. Additionally, each quadrotor independently runs its robot controller as specified in eqs.~\eqref{eq:ctrl_force_scalar} and \eqref{eq:ctrl_moment}, thus these components also maintain a computational complexity of $\mathcal{O}(1)$.

The dynamic force distribution involves two parts: nominal force distribution and human-aware force distribution. It requires linear mapping via matrix multiplication, as shown in eq.~\eqref{eq:destension}, with a complexity bound of $\mathcal{O}(n)$. 

Regarding human-aware force distribution, we propose two methods: an optimization-based method and a gradient-based method. We discuss them separately below:
\begin{enumerate}[label=\roman*)]
\item The optimization-based method employs the Sequential Least-Squares Quadratic Programming (SLSQP) solver, requiring $\mathcal{O}(a^2)$ storage and $\mathcal{O}(a^3)$ time, where $a = 3\times (n-6)$ represents the problem dimension.
\item The gradient-based method computes the cable tension force modifier through null space projection of a scaled gradient vector, optimizing eq.~\eqref{eq:costfunction}. The sum of $\mathcal{L}_2$-norm distances between each quadrotor and the human operator, combined with the closed-form solution for null space projection, leads to a computational complexity of $\mathcal{O}(n)$. This complexity is less than that of the optimization-based method, attributed to the utilization of closed-form solutions for both gradient computation and null space projection.
\end{enumerate}

\section{Conclusion and Future Works}~\label{sec:conclusion}
In this paper, we presented a human-aware human-robot collaborative transportation and manipulation approach considering a team of aerial robots with a cable-suspended payload. Our approach combines a novel control method that leverages system redundancy with a collaborative wrench estimator, enabling a human operator to interact in $6$ DoF with a rigid structure being transported by a team of aerial robots via cable. Additionally, the system can achieve secondary tasks like keeping a certain distance between the human operator and robots, or inter-robot separation by exploiting the additional system redundancy without affecting the quality or accuracy of the interactive experience. We demonstrated, through real-world experiments, our system's capabilities. The system can assist the human operator in manipulation tasks, as well as enable the human operator to effectively assist the load navigation, as demonstrated in the experiments.

In future research, we aim to expand our study into human-centric considerations, prioritizing metrics related to comfortness and acceptance of human operators. These elements are crucial in the domain of human-robot interaction. Additionally, we plan to develop safety methods to counteract unexpected human actions, such as sudden or forceful human physical inputs to the load that could lead to cable slack or actuator overload. This could ensure further robust operation under varied conditions.

In addition, we want to extend our framework to explicitly address collision avoidance between the cables and the human operator as well. Our current approach relies on the human's ability to navigate around the cables. However, by modeling the cables as convex polygons and incorporating them into the optimization process, we can develop a more comprehensive collision avoidance strategy that ensures human safety. Further developments also include the design of an onboard sensing mechanism. We plan to employ tension-measurement tools, onboard cameras, IMUs, and ESCs on each vehicle. Our goal is to achieve comprehensive onboard state estimation, therefore eliminating dependence on external motion capture systems. We also intend to investigate the impacts of state estimation and control delays, lags, and noise on system performance. Understanding these factors will enable us to improve our the robustness of our framework, enhancing interaction experience.% through the modeling and mitigation of these disturbances.

Finally, we would like to integrate a more advanced onboard perception module. It can empower the robot team to identify and navigate around complex hazardous spaces. This feature will enable autonomous obstacle avoidance maneuvers while leveraging the system's redundancy to maintain the intended payload trajectory without compromise. We  also envision employing deep learning techniques with robots' onboard cameras to analyze the human operator's posture, enhancing our human-aware force distribution strategy.

\bibliographystyle{IEEEtran}
\bibliography{references}

% Generated by IEEEtran.bst, version: 1.14 (2015/08/26)
\begin{thebibliography}{10}
\providecommand{\url}[1]{#1}
\csname url@samestyle\endcsname
\providecommand{\newblock}{\relax}
\providecommand{\bibinfo}[2]{#2}
\providecommand{\BIBentrySTDinterwordspacing}{\spaceskip=0pt\relax}
\providecommand{\BIBentryALTinterwordstretchfactor}{4}
\providecommand{\BIBentryALTinterwordspacing}{\spaceskip=\fontdimen2\font plus
\BIBentryALTinterwordstretchfactor\fontdimen3\font minus
  \fontdimen4\font\relax}
\providecommand{\BIBforeignlanguage}[2]{{%
\expandafter\ifx\csname l@#1\endcsname\relax
\typeout{** WARNING: IEEEtran.bst: No hyphenation pattern has been}%
\typeout{** loaded for the language `#1'. Using the pattern for}%
\typeout{** the default language instead.}%
\else
\language=\csname l@#1\endcsname
\fi
#2}}
\providecommand{\BIBdecl}{\relax}
\BIBdecl

\bibitem{Schwab_2017}
K.~Schwab, \emph{The fourth industrial revolution}.\hskip 1em plus 0.5em minus
  0.4em\relax London, England: Portfolio Penguin, 2017.

\bibitem{mansouriCooperativeCoveragePath2018}
S.~S. Mansouri, C.~Kanellakis, E.~Fresk, D.~Kominiak, and G.~Nikolakopoulos,
  ``Cooperative coverage path planning for visual inspection,'' \emph{Control
  Engineering Practice}, vol.~74, pp. 118--131, 2018.

\bibitem{shah2020multidronesurveys}
K.~Shah, G.~Ballard, A.~Schmidt, and M.~Schwager, ``Multidrone aerial surveys
  of penguin colonies in antarctica,'' \emph{Science Robotics}, vol.~5, no.~47,
  p. eabc3000, 2020.

\bibitem{giuseppe2018ijrr}
G.~Loianno, Y.~Mulgaonkar, C.~Brunner, D.~Ahuja, A.~Ramanandan, M.~Chari,
  S.~Diaz, and V.~Kumar, ``Autonomous flight and cooperative control for
  reconstruction using aerial robots powered by smartphones,'' \emph{The
  International Journal of Robotics Research}, vol.~37, no.~11, pp. 1341--1358,
  2018.

\bibitem{michael2012jfrcollaborativemapping}
N.~Michael, S.~Shen, K.~Mohta, Y.~Mulgaonkar, V.~Kumar, K.~Nagatani, Y.~Okada,
  S.~Kiribayashi, K.~Otake, K.~Yoshida, K.~Ohno, E.~Takeuchi, and S.~Tadokoro,
  ``Collaborative mapping of an earthquake-damaged building via ground and
  aerial robots,'' \emph{Journal of Field Robotics}, vol.~29, no.~5, pp.
  832--841, 2012.

\bibitem{cano2022jirs}
A.~E. {Jim{\'e}nez-Cano}, D.~Sanalitro, M.~Tognon, A.~Franchi, and
  J.~Cort{\'e}s, ``Precise {{Cable-Suspended Pick-and-Place}} with an {{Aerial
  Multi-robot System}},'' \emph{Journal of Intelligent \& Robotic Systems},
  vol. 105, no.~3, p.~68, 2022.

\bibitem{sanalitro2022ralindirectforce}
D.~Sanalitro, M.~Tognon, A.~J. Cano, J.~Cortés, and A.~Franchi, ``Indirect
  force control of a cable-suspended aerial multi-robot manipulator,''
  \emph{IEEE Robotics and Automation Letters}, vol.~7, no.~3, pp. 6726--6733,
  2022.

\bibitem{saskaSwarmDistributionDeployment2016}
M.~Saska, V.~Von{\'a}sek, J.~Chudoba, J.~Thomas, G.~Loianno, and V.~Kumar,
  ``Swarm distribution and deployment for cooperative surveillance by
  micro-aerial vehicles,'' \emph{Journal of Intelligent \& Robotic Systems},
  vol.~84, no.~1, pp. 469--492, 2016.

\bibitem{guanrui2021iser}
G.~Li and G.~Loianno, ``Design and experimental evaluation of distributed
  cooperative transportation of cable suspended payloads with micro aerial
  vehicles,'' in \emph{Experimental Robotics}, B.~Siciliano, C.~Laschi, and
  O.~Khatib, Eds.\hskip 1em plus 0.5em minus 0.4em\relax {Cham}: {Springer
  International Publishing}, 2021, pp. 28--36.

\bibitem{jackson2020raldistributedplanning}
B.~E. Jackson, T.~A. Howell, K.~Shah, M.~Schwager, and Z.~Manchester,
  ``Scalable cooperative transport of cable-suspended loads with {UAVs} using
  distributed trajectory optimization,'' \emph{IEEE Robotics and Automation
  Letters}, vol.~5, no.~2, pp. 3368--3374, 2020.

\bibitem{guanrui2021ral}
G.~Li, R.~Ge, and G.~Loianno, ``Cooperative transportation of cable suspended
  payloads with {MAV}s using monocular vision and inertial sensing,''
  \emph{IEEE Robotics and Automation Letters}, vol.~6, no.~3, pp. 5316--5323,
  2021.

\bibitem{ollero2021TRO}
A.~Ollero, M.~Tognon, A.~Suarez, D.~Lee, and A.~Franchi, ``Past, present, and
  future of aerial robotic manipulators,'' \emph{IEEE Transactions on
  Robotics}, vol.~38, no.~1, pp. 626--645, 2022.

\bibitem{afifi2023icuas}
A.~Afifi, G.~Corsini, Q.~Sable, Y.~Aboudorra, D.~Sidobre, and A.~Franchi,
  ``Physical human-aerial robot interaction and collaboration: Exploratory
  results and lessons learned,'' in \emph{International Conference on Unmanned
  Aircraft Systems (ICUAS)}, 2023, pp. 956--962.

\bibitem{tagliabue2019ijrr}
A.~Tagliabue, M.~Kamel, R.~Siegwart, and J.~Nieto, ``Robust collaborative
  object transportation using multiple {MAVs},'' \emph{The International
  Journal of Robotics Research}, vol.~38, no.~9, pp. 1020--1044, 2019.

\bibitem{LoiannoRAL2018}
G.~Loianno and V.~Kumar, ``Cooperative transportation using small quadrotors
  using monocular vision and inertial sensing,'' \emph{IEEE Robotics and
  Automation Letters}, vol.~3, no.~2, pp. 680--687, 2018.

\bibitem{Mellinger2013}
D.~Mellinger, M.~Shomin, N.~Michael, and V.~Kumar, \emph{Cooperative Grasping
  and Transport Using Multiple Quadrotors}.\hskip 1em plus 0.5em minus
  0.4em\relax Berlin, Heidelberg: Springer Berlin Heidelberg, 2013, pp.
  545--558.

\bibitem{thomas2014toward}
J.~Thomas, G.~Loianno, J.~Polin, K.~Sreenath, and V.~Kumar, ``Toward autonomous
  avian-inspired grasping for micro aerial vehicles,'' \emph{Bioinspiration \&
  biomimetics}, vol.~9, no.~2, p. 025010, 2014.

\bibitem{afifi2022icraphysicalinteractionaerialmanipulator}
A.~Afifi, M.~van Holland, and A.~Franchi, ``Toward physical human-robot
  interaction control with aerial manipulators: Compliance, redundancy
  resolution, and input limits,'' in \emph{IEEE International Conference on
  Robotics and Automation (ICRA)}, 2022, pp. 4855--4861.

\bibitem{corsini2022irosnmpchandover}
G.~Corsini, M.~Jacquet, H.~Das, A.~Afifi, D.~Sidobre, and A.~Franchi,
  ``Nonlinear model predictive control for human-robot handover with
  application to the aerial case,'' in \emph{{IEEE/RSJ International Conference
  on Intelligent Robots and Systems (IROS)}}, 2022.

\bibitem{tognon2021trophysicalinteractiontether}
M.~Tognon, R.~Alami, and B.~Siciliano, ``Physical human-robot interaction with
  a tethered aerial vehicle: Application to a force-based human guiding
  problem,'' \emph{IEEE Transactions on Robotics}, vol.~37, no.~3, pp.
  723--734, 2021.

\bibitem{kondak2020icrateleaerialmanipulator}
J.~Lee, R.~Balachandran, Y.~S. Sarkisov, M.~De~Stefano, A.~Coelho, K.~Shinde,
  M.~J. Kim, R.~Triebel, and K.~Kondak, ``Visual-inertial telepresence for
  aerial manipulation,'' in \emph{IEEE International Conference on Robotics and
  Automation (ICRA)}, 2020, pp. 1222--1229.

\bibitem{gengJGCD2020}
J.~Geng and J.~Langelaan, ``Cooperative transport of a slung load using
  load-leading control,'' \emph{Journal of Guidance, Control, and Dynamics},
  vol.~43, no.~7, pp. 1313--1331, 2020.

\bibitem{rastgoftar2019tcst}
H.~Rastgoftar and E.~M. Atkins, ``Cooperative aerial payload transport guided
  by an in situ human supervisor,'' \emph{IEEE Transactions on Control Systems
  Technology}, vol.~27, no.~4, pp. 1452--1467, 2019.

\bibitem{romano2022jgcd}
M.~Romano, A.~Ye, J.~Pye, and E.~Atkins, ``Cooperative multilift slung load
  transportation using haptic admittance control guidance,'' \emph{Journal of
  Guidance, Control, and Dynamics}, vol.~45, no.~10, pp. 1899--1912, 2022.

\bibitem{klausen2020passiveoutdoortransportation}
K.~Klausen, C.~Meissen, T.~I. Fossen, M.~Arcak, and T.~A. Johansen,
  ``Cooperative control for multirotors transporting an unknown suspended load
  under environmental disturbances,'' \emph{IEEE Transactions on Control
  Systems Technology}, vol.~28, no.~2, pp. 653--660, 2020.

\bibitem{Bernard2011jfr}
M.~Bernard, K.~Kondak, I.~Maza, and A.~Ollero, ``Autonomous transportation and
  deployment with aerial robots for search and rescue missions,'' \emph{Journal
  of Field Robotics}, vol.~28, no.~6, pp. 914--931, 2011.

\bibitem{tran2020itiianntransportation}
V.~P. Tran, F.~Santoso, M.~A. Garratt, and S.~G. Anavatti, ``Distributed
  artificial neural networks-based adaptive strictly negative imaginary
  formation controllers for unmanned aerial vehicles in time-varying
  environments,'' \emph{IEEE Transactions on Industrial Informatics}, vol.~17,
  no.~6, pp. 3910--3919, 2021.

\bibitem{zhang2021ralselftriggedtransportation}
X.~Zhang, F.~Zhang, P.~Huang, J.~Gao, H.~Yu, C.~Pei, and Y.~Zhang,
  ``Self-triggered based coordinate control with low communication for tethered
  multi-uav collaborative transportation,'' \emph{IEEE Robotics and Automation
  Letters}, vol.~6, no.~2, pp. 1559--1566, 2021.

\bibitem{LIU2021106673}
Y.~Liu, F.~Zhang, P.~Huang, and X.~Zhang, ``Analysis, planning and control for
  cooperative transportation of tethered multi-rotor {UAVs},'' \emph{Aerospace
  Science and Technology}, vol. 113, p. 106673, 2021.

\bibitem{lee2013cdccooperativepointmass}
T.~Lee, K.~Sreenath, and V.~Kumar, ``Geometric control of cooperating multiple
  quadrotor {UAVs} with a suspended payload,'' in \emph{52nd IEEE Conference on
  Decision and Control (CDC)}, 2013, pp. 5510--5515.

\bibitem{tagliabue2017icracollaborativetransportation}
A.~Tagliabue, M.~Kamel, S.~Verling, R.~Siegwart, and J.~Nieto, ``Collaborative
  transportation using mavs via passive force control,'' in \emph{IEEE
  International Conference on Robotics and Automation (ICRA)}, 2017, pp.
  5766--5773.

\bibitem{gassner2017icradynamiccollaboration}
M.~Gassner, T.~Cieslewski, and D.~Scaramuzza, ``Dynamic collaboration without
  communication: Vision-based cable-suspended load transport with two
  quadrotors,'' in \emph{IEEE International Conference on Robotics and
  Automation (ICRA)}, 2017, pp. 5196--5202.

\bibitem{rastgoftar2018icuas}
H.~Rastgoftar and E.~M. Atkins, ``Continuum deformation of a multiple
  quadcopter payload delivery team without inter-agent communication,'' in
  \emph{International Conference on Unmanned Aircraft Systems (ICUAS)}, 2018,
  pp. 539--548.

\bibitem{heng2022access}
H.~Xie, K.~Dong, and P.~Chirarattananon, ``Cooperative transport of a suspended
  payload via two aerial robots with inertial sensing,'' \emph{IEEE Access},
  vol.~10, pp. 81\,764--81\,776, 2022.

\bibitem{tognon2018passive}
M.~Tognon, C.~Gabellieri, L.~Pallottino, and A.~Franchi, ``Aerial
  co-manipulation with cables: The role of internal force for equilibria,
  stability, and passivity,'' \emph{IEEE Robotics and Automation Letters},
  vol.~3, no.~3, pp. 2577--2583, 2018.

\bibitem{lee2018tcst}
T.~Lee, ``Geometric control of quadrotor {UAVs} transporting a cable-suspended
  rigid body,'' \emph{IEEE Transactions on Control Systems Technology},
  vol.~26, no.~1, pp. 255--264, 2018.

\bibitem{wu2014geometric}
G.~Wu and K.~Sreenath, ``Geometric control of multiple quadrotors transporting
  a rigid-body load,'' in \emph{53rd IEEE Conference on Decision and Control
  (CDC)}, 2014, pp. 6141--6148.

\bibitem{fink2011ijrr}
J.~Fink, N.~Michael, S.~Kim, and V.~Kumar, ``Planning and control for
  cooperative manipulation and transportation with aerial robots,'' \emph{The
  International Journal of Robotics Research}, vol.~30, no.~3, pp. 324--334,
  2011.

\bibitem{Michael2011}
N.~Michael, J.~Fink, and V.~Kumar, ``Cooperative manipulation and
  transportation with aerial robots,'' \emph{Autonomous Robots}, vol.~30,
  no.~1, pp. 73--86, 2011.

\bibitem{sundin2022icradecentralizedmpctransportation}
R.~C. Sundin, P.~Roque, and D.~V. Dimarogonas, ``Decentralized model predictive
  control for equilibrium-based collaborative uav bar transportation,'' in
  \emph{IEEE International Conference on Robotics and Automation (ICRA)}, 2022,
  pp. 4915--4921.

\bibitem{tartaglione2017rasModelPredictiveControl}
G.~Tartaglione, E.~D’Amato, M.~Ariola, P.~S. Rossi, and T.~A. Johansen,
  ``\BIBforeignlanguage{en}{Model predictive control for a multi-body
  slung-load system},'' \emph{\BIBforeignlanguage{en}{Robotics and Autonomous
  Systems}}, vol.~92, pp. 1--11, 2017.

\bibitem{Leecdc2014}
T.~Lee, ``Geometric control of multiple quadrotor {UAVs} transporting a
  cable-suspended rigid body,'' in \emph{53rd IEEE Conference on Decision and
  Control (CDC)}, 2014, pp. 6155--6160.

\bibitem{geng2022jais}
J.~Geng, P.~Singla, and J.~W. Langelaan, ``Load-distribution-based trajectory
  planning and control for a multilift system,'' \emph{Journal of Aerospace
  Information Systems}, vol.~19, no.~5, pp. 366--381, 2022.

\bibitem{masone2016iros}
C.~Masone, H.~H. Bülthoff, and P.~Stegagno, ``Cooperative transportation of a
  payload using quadrotors: A reconfigurable cable-driven parallel robot,'' in
  \emph{IEEE/RSJ International Conference on Intelligent Robots and Systems
  (IROS)}, 2016, pp. 1623--1630.

\bibitem{bulka2022outdoortransportation}
E.~Bulka, C.~He, J.~Wehbeh, and I.~Sharf, ``Experiments on collaborative
  transport of cable-suspended payload with quadrotor {UAVs},'' in
  \emph{International Conference on Unmanned Aircraft Systems (ICUAS)}, 2022,
  pp. 1465--1473.

\bibitem{spaa2020icra}
L.~v. der Spaa, M.~Gienger, T.~Bates, and J.~Kober, ``Predicting and optimizing
  ergonomics in physical human-robot cooperation tasks,'' in \emph{{IEEE
  International Conference on Robotics and Automation (ICRA)}}, 2020, pp.
  1799--1805.

\bibitem{Stuckler_Behnke_2011}
J.~Stuckler and S.~Behnke, ``\BIBforeignlanguage{en}{Following human guidance
  to cooperatively carry a large object},'' in
  \emph{\BIBforeignlanguage{en}{11th IEEE-RAS International Conference on
  Humanoid Robots}}, Bled, Slovenia, 2011, p. 218–223.

\bibitem{gienger2018iros}
M.~Gienger, D.~Ruiken, T.~Bates, M.~Regaieg, M.~MeiBner, J.~Kober, P.~Seiwald,
  and A.-C. Hildebrandt, ``Human-robot cooperative object manipulation with
  contact changes,'' in \emph{{IEEE/RSJ International Conference on Intelligent
  Robots and Systems (IROS)}}, 2018, pp. 1354--1360.

\bibitem{Sheng_Thobbi_Gu_2015}
W.~Sheng, A.~Thobbi, and Y.~Gu, ``An integrated framework for human–robot
  collaborative manipulation,'' \emph{IEEE Transactions on Cybernetics},
  vol.~45, no.~10, p. 2030–2041, 2015.

\bibitem{Augugliaro_2013_ecc}
F.~Augugliaro and R.~D’Andrea, ``Admittance control for physical
  human-quadrocopter interaction,'' in \emph{European Control Conference
  (ECC)}, 2013, p. 1805–1810.

\bibitem{SieberIROS2015}
D.~Sieber, S.~Musić, and S.~Hirche, ``Multi-robot manipulation controlled by a
  human with haptic feedback,'' in \emph{IEEE/RSJ International Conference on
  Intelligent Robots and Systems (IROS)}, 2015, p. 2440–2446.

\bibitem{Yashin_Trinitatova_Agishev_Ibrahimov_Tsetserukou_2019}
G.~A. Yashin, D.~Trinitatova, R.~T. Agishev, R.~Ibrahimov, and D.~Tsetserukou,
  ``Aerovr: Virtual reality-based teleoperation with tactile feedback for
  aerial manipulation,'' in \emph{International Conference on Advanced Robotics
  (ICAR)}, 2019, p. 767–772.

\bibitem{Sachidanandam_Honarvar_Diaz-Mercado_2022}
S.~O. Sachidanandam, S.~Honarvar, and Y.~Diaz-Mercado, ``Effectiveness of
  augmented reality for human swarm interactions,'' in \emph{IEEE International
  Conference on Robotics and Automation (ICRA)}, 2022, p. 11258–11264.

\bibitem{Elwin_Strong_Freeman_Lynch_2023}
M.~L. Elwin, B.~Strong, R.~A. Freeman, and K.~M. Lynch, ``Human-multirobot
  collaborative mobile manipulation: The omnid mocobots,'' \emph{IEEE Robotics
  and Automation Letters}, vol.~8, no.~1, p. 376–383, 2023.

\bibitem{Sirintuna_Ozdamar_Ajoudani_2023}
D.~Sirintuna, I.~Ozdamar, and A.~Ajoudani, ``Carrying the uncarriable: a
  deformation-agnostic and human-cooperative framework for unwieldy objects
  using multiple robots,'' in \emph{{IEEE International Conference on Robotics
  and Automation (ICRA)}}, 2023, pp. 4915--4921.

\bibitem{Carey_Werfel_2022}
N.~E. Carey and J.~Werfel, ``\BIBforeignlanguage{en}{A force-mediated
  controller for cooperative object manipulation with independent autonomous
  robots},'' in \emph{\BIBforeignlanguage{en}{International Symposium on
  Distributed Autonomous Robotic Systems}}, 2022.

\bibitem{guanrui2022rotortm}
G.~Li, X.~Liu, and G.~Loianno, ``Rotortm: A flexible simulator for aerial
  transportation and manipulation,'' \emph{IEEE Transactions on Robotics}, pp.
  1--20, 2023.

\bibitem{guanrui2023iros}
G.~Li and G.~Loianno, ``Nonlinear model predictive control for cooperative
  transportation and manipulation of cable suspended payloads with multiple
  quadrotors,'' in \emph{{IEEE/RSJ International Conference on Intelligent
  Robots and Systems (IROS)}}, 2023, pp. 1--7.

\bibitem{siciliano_robotics_2009}
B.~Siciliano, L.~Sciavicco, L.~Villani, and G.~Oriolo, \emph{Robotics
  Modelling, Planning and Control}.\hskip 1em plus 0.5em minus 0.4em\relax
  Springer London, 2009.

\bibitem{Bertsekas/99}
D.~Bertsekas, \emph{Nonlinear Programming}.\hskip 1em plus 0.5em minus
  0.4em\relax Athena Scientific, 1999.

\bibitem{dieter1994acmnlsqp}
D.~Kraft, ``Algorithm 733: Tomp–fortran modules for optimal control
  calculations,'' \emph{ACM Trans. Math. Softw.}, vol.~20, no.~3, p. 262–281,
  1994.

\bibitem{johnson2011nlopt}
\BIBentryALTinterwordspacing
S.~G. Johnson, \emph{The NLopt nonlinear-optimization package}, 2011. [Online].
  Available: \url{http://ab-initio.mit.edu/nlopt}
\BIBentrySTDinterwordspacing

\bibitem{mellinger_snap_icra_2011}
D.~Mellinger and V.~Kumar, ``Minimum snap trajectory generation and control for
  quadrotors,'' in \emph{{IEEE International Conference on Robotics and
  Automation (ICRA)}}, 2011, pp. 2520--2525.

\bibitem{sola2017quaternion}
J.~Sola, ``Quaternion kinematics for the error-state kalman filter,''
  \emph{arXiv preprint arXiv:1711.02508}, 2017.

\bibitem{thrun_probabilistic_robotics}
S.~Thrun, W.~Burgard, and D.~Fox, \emph{Probabilistic Robotics}.\hskip 1em plus
  0.5em minus 0.4em\relax MIT Press, 2005.

\bibitem{LoiannoRAL2017}
G.~Loianno, C.~Brunner, G.~McGrath, and V.~Kumar, ``Estimation, control, and
  planning for aggressive flight with a small quadrotor with a single camera
  and imu,'' \emph{IEEE Robotics and Automation Letters}, vol.~2, no.~2, pp.
  404--411, 2017.

\bibitem{rennips2015fasterrcnh}
S.~Ren, K.~He, R.~Girshick, and J.~Sun, ``Faster {R-CNN}: Towards real-time
  object detection with region proposal networks,'' in \emph{Advances in Neural
  Information Processing Systems}, C.~Cortes, N.~Lawrence, D.~Lee, M.~Sugiyama,
  and R.~Garnett, Eds., vol.~28.\hskip 1em plus 0.5em minus 0.4em\relax Curran
  Associates, Inc., 2015.

\end{thebibliography}
\begin{IEEEbiography}[{\includegraphics[width=1in,height=1.5in,clip,keepaspectratio]{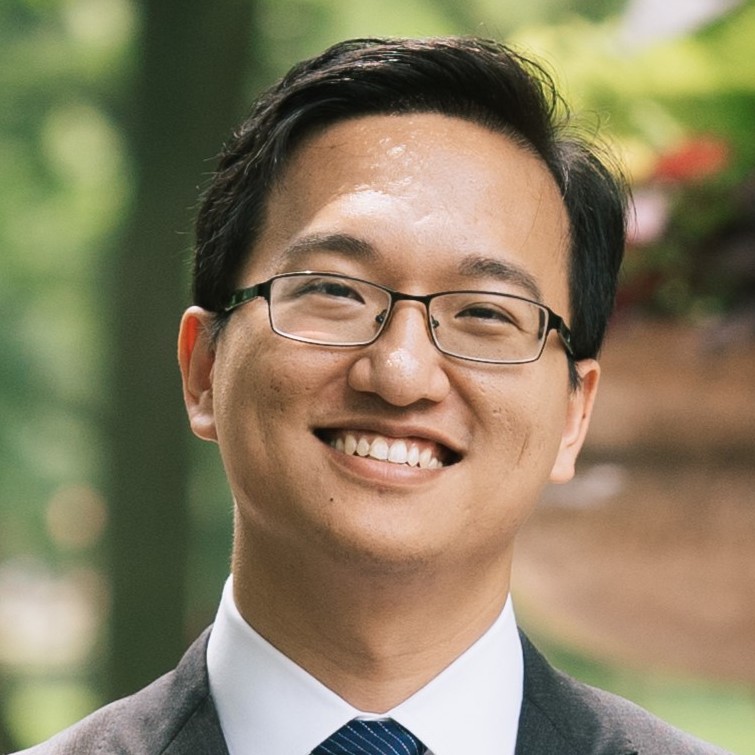}}]{Guanrui Li} is an Assistant Professor at the Worceter Polytechnic Institute (WPI), USA and director of the Aerial-robot Control and Perception Lab (ACP Lab) at WPI. He earned his Ph.D. in Electrical and Computer Engineering at New York University, USA, with a focus on robotics and aerial systems. He obtained his Bachelor degree in Theoretical and Applied Mechanics from Sun Yat-sen University, where he was recognized as an Honors Undergraduate, and his Master degree in Robotics from the GRASP Lab at the University of Pennsylvania. His research is centered on the dynamics, planning, and control of robotics systems, with applications in aerial transportation and manipulation, as well as human-robot collaboration. Guanrui has received several notable recognitions, including the NSF CPS Rising Stars in 2023, the  Outstanding Deployed System Paper Award finalist at 2022 IEEE ICRA, and the 2022 Dante Youla Award for Graduate Research Excellence at NYU. He has an extensive publication record in top-tier robotics conferences and journals like ICRA, RA-L, and T-RO, and his work has garnered attention in various media, including IEEE Spectrum and the Discovery Channel.
\end{IEEEbiography}

\begin{IEEEbiography}[{\includegraphics[width=1in,height=1.5in,clip,keepaspectratio]{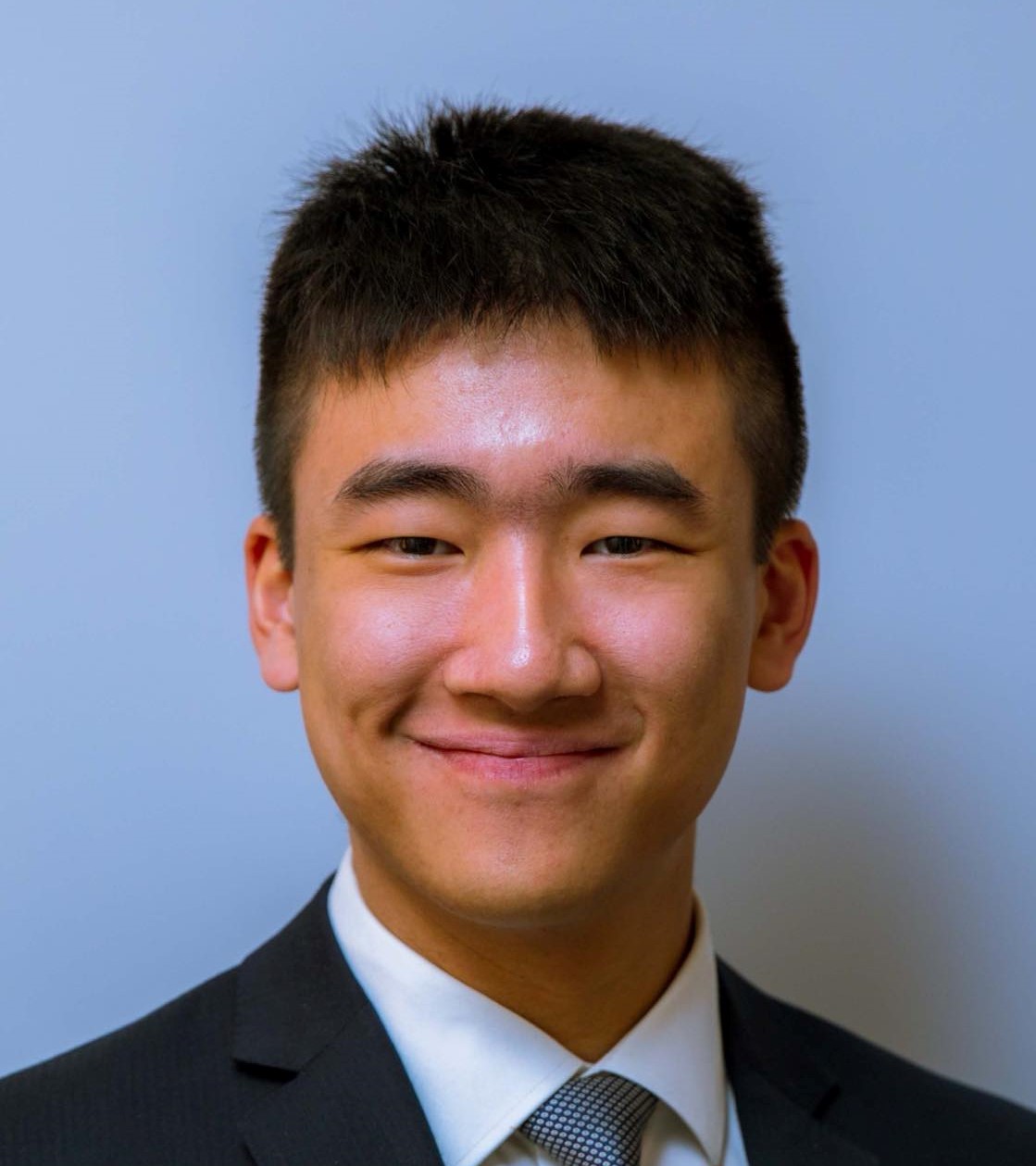}}]{Xinyang Liu}
Xinyang Liu was borned in Beijing, China, in 1999. He received his B.S. degree in mechanical engineering with double minors in robotics and computer science from New York University, New York, NY, in 2022, where he was named a University Honor Scholar and was a member of Tau Beta Pi. He is currently in his final year pursuing an M.S. degree in Aeronautics and Astronautics at Stanford University, Stanford, CA. His research interests include control theory, robotics, human-robot interactions, and autonomous systems.
\end{IEEEbiography}

\begin{IEEEbiography}[{\includegraphics[width=1in,height=1.5in,clip,keepaspectratio]{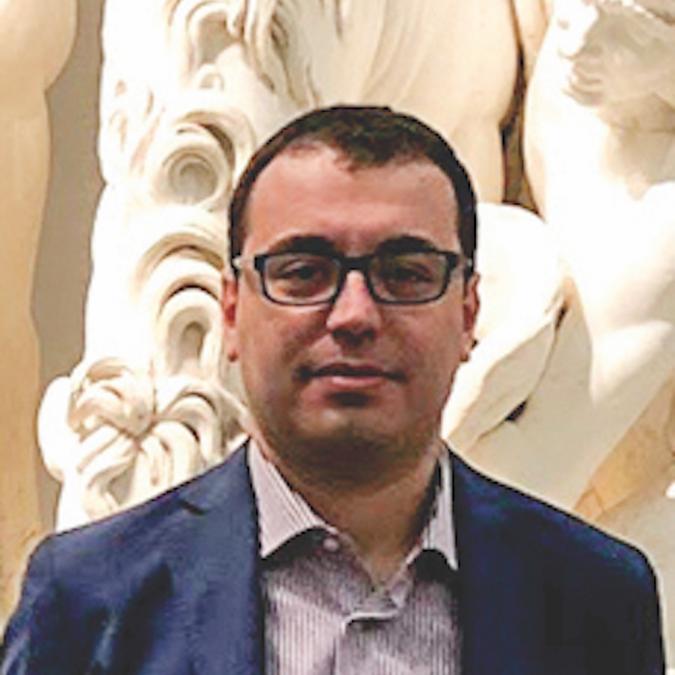}}]{Giuseppe Loianno} is an assistant professor at the New York University, USA and director of the Agile Robotics and Perception Lab (https://wp.nyu.e
du/arpl/) working on autonomous robots. He received a Ph.D. in robotics from University of Naples ”Federico II”, Italy in 2014. Prior joining NYU, he was post-doctoral researcher, research scientist and
team leader at the GRASP Lab at the University of
Pennsylvania in Philadelphia, USA. Dr. Loianno has
published more than 70 conference papers, journal
papers, and book chapters. His research interests
include perception, learning, and control for autonomous robots. He received
the NSF CAREER Award in 2022 and DARPA Young Faculty Award in
2022. He is recipient of the IROS Toshio Fukuda Young Professional Award
in 2022, Conference Editorial Board Best Associate Editor Award at ICRA
2022, Best Reviewer Award at ICRA 2016, and he was selected as Rising
Star in AI from KAUST in 2023. He is also currently the co-chair of the
IEEE RAS Technical Committee on Aerial Robotics and Unmanned Aerial
Vehicles. He was the the general chair of the IEEE International Symposium
on Safety, Security and Rescue Robotics (SSRR) in 2021 as well as program
chair in 2019, 2020, and 2022. His work has been featured in a large number
of renowned international news and magazines.

\end{IEEEbiography}
\vfill
%\biboptions{sort&compress}
%\input{10-old.tex}
\end{document}